\documentclass[pdflatex,sn-mathphys-num]{sn-jnl}

\usepackage{graphicx}%
\usepackage{multirow}%
\usepackage{amsmath,amssymb,amsfonts}%
\usepackage{amsthm}%
\usepackage{mathrsfs}%
\usepackage[title]{appendix}%
\usepackage{xcolor}%
\usepackage{textcomp}%
\usepackage{manyfoot}%
\usepackage{booktabs}%
\usepackage{algorithm}%
\usepackage{algorithmicx}%
\usepackage{algpseudocode}%
\usepackage{listings}%

\usepackage{booktabs,makecell}
\usepackage{algorithm}
\usepackage{graphicx,subcaption}

\theoremstyle{thmstyleone}%
\newtheorem{theorem}{Theorem}
\newtheorem{proposition}[theorem]{Proposition}%

\theoremstyle{thmstyletwo}%

\theoremstyle{thmstylethree}%

\begin{document}

\title[Article Title]{A decoupled alignment kernel for peptide membrane permeability predictions}

\author*[1,3]{\fnm{Ali} \sur{Amirahmadi}}\email{ali.amirahmadi@hh.se}

\author[2,3]{\fnm{G\"{o}k\c{c}e} \sur{Geylan}}\email{gokce.geylan@astrazeneca.com}

\author[4]{\fnm{Leonardo} \sur{De Maria}}\email{leonardo.demaria@astrazeneca.com}

\author[1,5]{\fnm{Farzaneh} \sur{Etminani}}\email{farzaneh.etminani@hh.se}

\author[1,6]{\fnm{Mattias} \sur{Ohlsson}}\email{mattias.ohlsson@hh.se}

\author[3]{\fnm{Alessandro} \sur{Tibo}}\email{alessandro.tibo@astrazeneca.com}

\affil*[1]{\orgdiv{Center for Applied Intelligent Systems Research in Health}, \orgname{Halmstad University}, \orgaddress{\street{Kristian IV:s v\"{a}g 3}, \city{Halmstad}, \postcode{30118}, \country{Sweden}}}

\affil[2]{\orgdiv{Division of Systems and Synthetic Biology, Department of Life Sciences}, \orgname{Chalmers University of Technology}, \orgaddress{\street{Kemig\aa{}rden 1}, \city{Gothenburg}, \postcode{41296}, \country{Sweden}}}

\affil[3]{\orgdiv{Molecular AI, Discovery Sciences, R\&D}, \orgname{AstraZeneca}, \orgaddress{\street{Pepparedsleden 1}, \city{M\"{o}lndal}, \postcode{43183}, \country{Sweden}}}

\affil[4]{\orgdiv{Medicinal Chemistry, Research and Early Development, Respiratory \& Immunology, BioPharmaceuticals R\&D}, \orgname{AstraZeneca}, \orgaddress{\street{Pepparedsleden 1}, \city{M\"{o}lndal}, \postcode{43183}, \country{Sweden}}}

\affil[5]{\orgdiv{Department of Research and Development (FoU)}, \orgname{Region Halland}, \orgaddress{\street{Södra v\"{a}gen 9}, \city{Halmstad}, \postcode{30118}, \country{Sweden}}}

\affil[6]{\orgdiv{Centre for Environmental and Climate Science, Computational Science for Health and Environment}, \orgname{Lund University}, \orgaddress{\street{Kontaktv\"{a}gen 10}, \city{Lund}, \postcode{22362}, \country{Sweden}}}



\abstract{
Cyclic peptides are promising modalities for targeting intracellular sites; however, cell-membrane permeability remains a key bottleneck, exacerbated by limited public data and the need for well-calibrated uncertainty. Instead of relying on data-eager complex deep learning architecture, we propose a monomer-aware decoupled global alignment kernel (MD-GAK), which couples chemically meaningful residue–residue similarity with sequence alignment while decoupling local matches from gap penalties. MD-GAK is a relatively simple kernel. To further demonstrate the robustness of our framework, we also introduce a variant, PMD-GAK, which incorporates a triangular positional prior. As we will show in the experimental section, PMD-GAK can sometimes offer additional advantages over MD-GAK, particularly in reducing calibration errors. 
Since our focus is on uncertainty estimation, we use Gaussian Processes as the predictive model, as both MD-GAK and PMD-GAK can be directly applied within this framework. We demonstrate the effectiveness of our methods through an extensive set of experiments, comparing our fully reproducible approach against state-of-the-art transformer-based models, and show that it outperforms them across all metrics.
}




\maketitle

\section{Introduction}\label{sec1}

Cyclic peptides have re-emerged as a compelling modality for intracellular targets thanks to their high affinity and selectivity, yet their \emph{cell membrane permeability} remains a central bottleneck for discovery~\citep{li2023cycpeptmpdb,geylan2024methodology}. Recent resources such as Cyclic Peptide Membrane Permeability Database (CycPeptMPDB) \citep{li2023cycpeptmpdb} aggregate permeability measurements from the literature and patents across multiple assays (notably Parallel artificial membrane permeability assay (PAMPA)), providing a large and well-annotated dataset for systematic modeling~\citep{li2023cycpeptmpdb}. However, these datasets are still relatively small by modern machine learning (ML) standards, heterogeneous in experimental provenance, and structurally biased, which complicates out-of-domain generalization and model calibration~\citep{geylan2024methodology,liu2025systematic}. These challenges underscore the need for methods that can (i) encode peptide-specific structure, (ii) work well in the small–to–medium data regime, and (iii) provide reliable uncertainty quantification.

Gaussian processes (GPs) are natural candidates for such settings: they deliver calibrated probabilistic predictions with an inductive bias entirely controlled by a \emph{kernel}~\citep{cuturi2011fast,perez2024gaussian}. The kernel choice is therefore pivotal. For small molecules, the Tanimoto (Jaccard) kernel on circular fingerprints (e.g., RDKit Morgan) has been highly successful. These fingerprints encode rich {local} topological environments via circular neighborhoods. However, when a cyclic peptide is represented as a single molecular graph, circular fingerprints are inherently order-agnostic with respect to the peptide’s monomer sequence \citep{landrum2016rdkit, rogers2010extended}: they encode local topological environments but do not preserve the explicit residue order or enable residue-to-residue alignment. Since permeability depends not only on which building blocks are present but also on where they occur and which neighbors they have \citep{wang2021effect, hosono2023amide}, this motivates kernels that pair chemistry-aware monomer similarities with sequence alignment, capturing both composition and arrangement.

Recent benchmarking on cyclic peptide permeability underscores that representation and split strategy (e.g., scaffold versus random splits) can dominate performance and robustness trends, and that calibrated uncertainty is as important as raw discrimination~\citep{liu2025systematic}. The work of parallel methodology also emphasizes careful data curation, duplicate control, and applicability domain-aware evaluation to avoid optimistic estimates under nearly duplicate leakage~\citep{geylan2024methodology}.

To retain peptide-specific structure, one can move beyond features from traditional fingerprint encodings that summarize each peptide as an order-invariant collection of local substructures and instead employ sequence- and graph-aware representations. Circular fingerprints such as Morgan/ECFP encode rich local chemistry and remain strong baselines for small-molecule QSAR; however, when a single global fingerprint is used as the only feature for a peptide, structural information (e.g. order and long-range topology) \cite{meyer2025reverse}, and subtle stereochemical rearrangements \cite{tahil2024stereoisomers} are largely missed. As a result, traditional classifiers such as SVMs or random forests often show limited discriminative power for cyclic peptide permeability when compared with modern graph-based models \cite{liu2025systematic}. On the sequence side, a large body of work now treats molecules explicitly as strings or token sequences. Recurrent and transformer-based chemical language models (e.g., LSTMs and ChemBERTa) operate on SMILES and have demonstrated that sequence-aware encoders can capture both local and more global structural patterns that are not easily accessible to a single fingerprint vector \citep{chithrananda2020chemberta}. 
Within kernel methods, \emph{Global Alignment} (GA) kernels provide a principled way to exploit sequence order: they replace the hard minimum of Dynamic Time Warping (DTW)~\citep{senin2008dynamic} with a soft sum over all monotone alignments, yielding positive-definite similarities suitable for kernel machines and Gaussian processes \citep{cuturi2007kernel}. "Triangular" GA variants introduce a kernel that both encodes positional priors and reduces computation, maintaining positive definiteness~\citep{cuturi2011fast}.

Graph-aware methods push this idea further by working directly on the molecular graph. Message-Passing Neural Networks (MPNNs) unify a broad family of graph convolutional architectures into a common message-passing framework, achieving state-of-the-art performance on several quantum-chemical benchmarks by learning task-specific node and edge embeddings from molecular graphs \citep{gilmer2017neural} . AttentiveFP builds on this paradigm by introducing a graph attention mechanism that adaptively reweights neighbors: it not only propagates information along local bonds but also learns nonlocal intramolecular interactions and “hidden’’ edges that are most relevant for a given property, while still respecting the underlying molecular topology \citep{xiong2019pushing}. Directed MPNNs (DMPNNs) refine message passing even further by operating on directed edges and improve information flow \citep{yang2019analyzing}. Recent systematic benchmarking on cyclic peptide membrane permeability demonstrates that graph-based models such as AttentiveFP, MPNN, and DMPNN are among the top-performing approaches, clearly outperforming classical SVM and random-forest models built solely on fingerprint descriptors.\citep{liu2025systematic}.

In this work, we take a complementary approach that combines the strengths of rich local chemistry of molecular fingerprints with sequence-aware topology. We introduce a \emph{monomer decoupled global alignment kernel} (MD-GAK) for cyclic peptides. Each peptide is represented as an ordered sequence of monomer units. Our key design choice is to treat each monomer as a small molecule and encode it with a Morgan fingerprint, while explicitly retaining both the sequence order and the chirality of the monomers.  Chirality is critical for Peptide permeability as minimal changes—as simple as a single stereochemical inversion—can rewire intramolecular networks and conformational ensembles and thus shift membrane permeability~\cite{li2023cycpeptmpdb, wang2021effect, hosono2023amide}. 
By feeding the resulting sequence of monomer-level fingerprints into a GA kernel, MD-GAK captures sequential topology while leveraging chemically rich local descriptors, yielding a structured similarity measure that is directly usable within kernel machines and Gaussian process models for cyclic peptide permeability prediction.

While we consider simple kernels on monomers in this work, our approach will, in principle, allow us to use any complex kernel (e.g., based on neural networks, transformers, ...) for small molecules for which we have access to billions of data points, in contrast to peptides, where the number of structures is limited to a few thousand. 
This decoupling lets us compare peptides via \emph{monomer–monomer} chemistry—where small-molecule tooling, fingerprints, and kernels are mature. We score local matches using a Tanimoto kernel on monomer fingerprints and then aggregate over all residue alignments with a GA-style dynamic program to form the peptide–peptide kernel. We further propose a \emph{position-aware} variant (PMD-GAK) that adds a triangular positional prior, reducing computational cost and discouraging implausible warpings (insertions and deletions in alignment). Embedded in a GP classifier, these kernels combine chemically grounded local similarity with flexible sequence alignment and yield calibrated predictive uncertainties—leveraging the strengths of small-molecule representations while capturing the sequence organization unique to peptides.

\textbf{Empirical scope.} Using CycPeptMPDB, we assess our approach in two complementary regimes: (i) an {applicability-domain–aware} nested cross-validation with label- and canonical-group–stratified folds to curb leakage and probe robustness~\citep{geylan2024methodology}; and (ii) a {length-focused} subset (6-, 7-, and 10-mers) evaluated under random and scaffold splits, following recent benchmarks~\citep{liu2025systematic}. Across both settings we compare against strong baselines. We observe consistent gains in discrimination (ACC/F1/ROC–AUC) and improved probabilistic calibration (Brier/ECE), with the position-aware variant particularly enhancing calibration.

\textbf{Contributions.}  
(1) A peptide-specific \emph{monomer decoupled global alignment Kernel} that aligns sequences of monomer-level molecular fingerprints, bridging local chemical similarity and sequence order.  
(2) A \emph{position-aware} triangular variant (PMD-GAK) that encodes positional priors and improves calibration while retaining positive-definite structure.  
(3) A thorough evaluation on CycPeptMPDB under leakage-aware protocols and length-focused splits, showing that alignment-aware GPs  exceed strong baselines and yield improved probabilistic calibration.  
(4) A theoretical construction (Appendix~\ref{append:proof}) proving the positive semidefiniteness of our kernel via a rational/convolution kernel argument, and a discussion of its relation to classical GA/Triangular-GA kernels and modern graph GP kernels~\citep{perez2024gaussian}.

Overall, our results indicate that \emph{monomer-aware alignment} is a practical and principled route to modeling cyclic peptide permeability: it captures sequence-sensitive chemistry, behaves well in the low-data regime typical of permeability measurements, and integrates seamlessly with GP-based uncertainty quantification---a combination that directly addresses the reliability concerns raised in recent methodological studies~\citep{geylan2024methodology,liu2025systematic}.

\section{Methods}
We cast peptide–peptide comparison as a positive-definite global-alignment kernel at the monomer level and use it as the covariance of a Gaussian process (GP). This connects chemically meaningful local similarity with sequence order, while retaining closed-form GP inference and calibrated uncertainties.


\subsection{Gaussian Process}
A Gaussian process (GP) is a distribution over functions \(f:\mathbb{R}^n\!\to\!\mathbb{R}\) such that any finite collection has a joint Gaussian distribution:
\[
  f(\cdot)\sim \mathcal{GP}\big(m(\cdot),\,k(\cdot,\cdot)\big),
\]
with mean function \(m\) and positive-definite covariance (kernel) \(k\). For a dataset $\mathcal{D} = \{ (x_i \in \mathbb{R}^n, y_i\in \mathbb{R})\}_{i=1}^{N}$, we represent all the inputs in a matrix $X$ of size $N\times n$ and the outputs in a vector $\mathbf{y}$ of size $N$.
The induced prior over function values is
\[
  \mathbf{f}:=f(X)\sim \mathcal{N}\!\big(m(X),\,K_{XX}\big),\quad (K_{XX})_{ij}=k(x_i,x_j).
\]
In regression tasks, we assume that the labels are noisy, i.e. 
\(y_i = f(x_i) + \varepsilon_i\) with 
\(\varepsilon_i \sim \mathcal{N}(0,\eta^2)\), 
where \(\eta > 0\) denotes the observation noise standard deviation.


\paragraph{Joint prior over train/test (regression).}
Let \(X_*\) denote test inputs and \(\mathbf{f}_*:=f(X_*)\). The joint prior over observed targets \(\mathbf{y}\) and test function values is
\[
\begin{bmatrix}
\mathbf{y}\\[2pt]
\mathbf{f}_*
\end{bmatrix}
\sim
\mathcal{N}\!\left(
\begin{bmatrix}
m(X)\\[2pt]
m(X_*)
\end{bmatrix},
\begin{bmatrix}
K_{XX}+\eta^2 I & K_{X X_*}\\
K_{X_* X} & K_{X_* X_*}
\end{bmatrix}
\right),
\]
with \(K_{X X_*}=k(X,X_*)\), \(K_{X_* X}=K_{X X_*}^{\top}\), and \(K_{X_* X_*}=k(X_*,X_*)\).

\paragraph{Posterior (regression).}
Conditioning the GP on \((X,\mathbf{y})\), we obtain a Gaussian posterior for \(\mathbf{f}_*\):
\[
  \bar{m}_* \;=\; m(X_*) \;+\; K_{X_* X}\big(K_{XX}+\eta^2 I\big)^{-1}\!\big(\mathbf{y}-m(X)\big),
\]
\[
  \bar{\Sigma}_* \;=\; K_{X_* X_*} \;-\; K_{X_* X}\big(K_{XX}+\eta^2 I\big)^{-1} K_{X X_*}.
\]
The posterior mean provides point predictions and \(\bar{\Sigma}_*\) quantifies predictive uncertainty.

\begin{figure}[ht]
\centering
\begin{minipage}{0.48\linewidth}
  \centering
  \includegraphics[width=\linewidth]{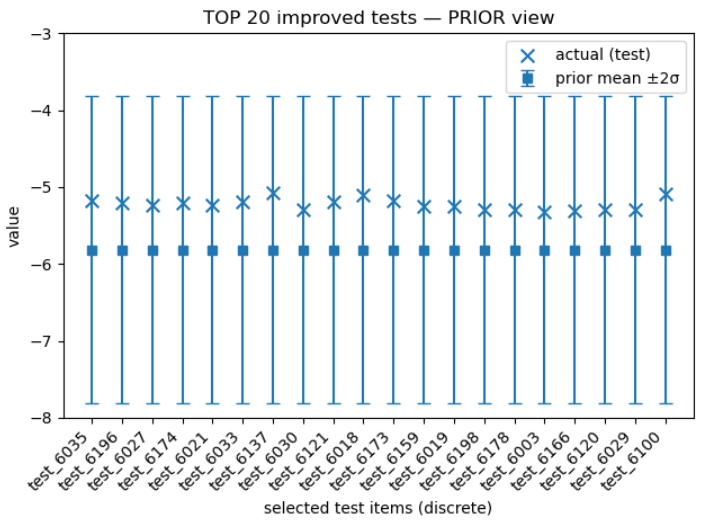}
  \vspace{2pt}
  \small (a) Prior draws
\end{minipage}\hfill
\begin{minipage}{0.48\linewidth}
  \centering
  \includegraphics[width=\linewidth]{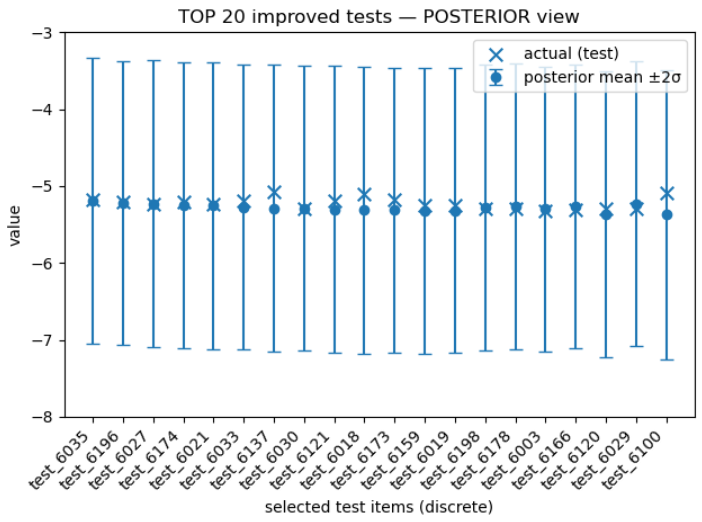}
  \vspace{2pt}
  \small (b) Posterior predictive
\end{minipage}
\caption{GP prior and posterior for PAMPA prediction using the proposed kernel. The data conditioning contracts the posterior uncertainty and shifts the mean toward test samples observations.}
\label{fig:two-images2}
\end{figure}

\paragraph{Binary GP classification (Laplace).}
For binary labels \(y_i\in\{0,1\}\) we place a zero-mean GP prior on latent logits,
\(\mathbf{f}\sim\mathcal{N}(0,K)\), and use a logistic likelihood
\(p(y_i=1\mid f_i)=\Psi(f_i)\) with \(\Psi(z)=1/(1+e^{-z})\).
The approximate posterior over \(\mathbf{f}\) is \(\mathcal{N}(\hat{\mathbf{f}},\,\Sigma)\) with
\(\Sigma=(K^{-1}+W)^{-1}\), where \(\hat{\mathbf{f}}\) is obtained using a Laplace approximation~\cite{rasmussen2003gaussian, williams2006gaussian} and \(W=\mathrm{diag}\!\big(\Psi(\hat{\mathbf{f}})\odot(1-\Psi(\hat{\mathbf{f}}))\big)\).
For a test input \(x_*\) with \(k_*:=k(X,x_*)\) and \(k_{**}:=k(x_*,x_*)\),
\[
\mu_{f_*}=k_*^{\top}\,K^{-1}\,\hat{\mathbf{f}}, \quad \sigma_{f_*}^2=k_{**}\;-\;k_*^{\top}\,\big(K+W^{-1}\big)^{-1}k_*,
\]
\[
p(y_*=1 \, | \, X, y, x_*)  = \int \sigma(f_*)\,\mathcal{N}(f_*;\mu_{f_*},\sigma_{f_*}^2)\,df_*\,.
\]

\subsubsection{Kernel methods}
In Gaussian processes (GPs), specifying a covariance (kernel) \(k\) determines a prior over functions and thereby encodes the inductive bias of the model. The predictive behavior of a GP depends critically on this choice: different kernels express different notions of similarity (e.g. smoothness, invariance, compositional structure) and thus suit different input types (e.g. strings, graphs, molecules).

Kernel methods specify a positive-definite kernel \(k(x,x')\) that represents an inner product in a (possibly infinite-dimensional) feature space \(\mathcal{H}\), i.e.
\[
k(x,x') \;=\; \langle \phi(x), \phi(x') \rangle_{\mathcal{H}}
\]
for some (typically implicit) feature map \(\phi\). 
A positive definite kernel $k:\mathcal{X}\times\mathcal{X}\to\mathbb{R}$ 
on an input space $\mathcal{X}$
implicitly defines an RKHS $\mathcal{H}_k$ in which linear algorithms operate via Gram matrices $K_{ij}=k(x_i,x_j)$ (the ``kernel trick'') \citep{scholkopf2002learning,shawe2004kernel}. In GPs, the kernel plays the role of a covariance function that encodes prior smoothness and inductive bias; posterior predictions follow in closed form given $K$ and a likelihood \citep{williams2006gaussian}. Consequently, the modelling power of a GP hinges on choosing a kernel tailored to the structure of the inputs (e.g. strings, graphs, molecules), rather than generic Euclidean distances.

\subsection{Monomer Global Alignment Kernel}\label{sec:m-gak}

\subsubsection{Global alignment}\label{sec:ga-background}
Given two sequences $A=(a_1,\ldots,a_n)$ and $B=(b_1,\ldots,b_m)$ of lengths $n$ and $m$, respectively over an alphabet $\mathcal{A}$, a (global) alignment is a pair of nondecreasing index paths $(\pi_1,\pi_2)$ with unit steps $(1,0)$, $(0,1)$, or $(1,1)$ that map $A$ to $B$ through substitutions and gaps. Dynamic programming (DP) aggregates local substitution scores and gap operations from boundary conditions to the terminal cell, in $O(nm)$ time \citep{cuturi2011fast}. 

While the Dynamic Time Warping (DTW) distance is a de-facto baseline for sequence comparison, it is \emph{not} a metric and is known to be not negative definite~\citep{cuturi2011fast}; consequently, kernels derived directly from DTW (e.g., $\exp(-\gamma\,\mathrm{DTW})$) are not guaranteed positive definite and can break RKHS-based learning. Global Alignment (GA) kernels address this by replacing the hard minimum in DTW with a sum over all monotone alignments (a soft-min), yielding a similarity that can be made positive definite under mild conditions on the local kernel  and gap weighting~\citep{cuturi2007kernel,cuturi2011fast}. GA kernels retain DTW’s $O(nm)$ complexity but, unlike DTW’s single best path, they summarize the entire ensemble of alignments costs, which is often more informative for learning~\citep[§2.2]{cuturi2011fast}.

\subsubsection{Global alignment kernels}\label{sec:ga-kernels}
Let $\kappa:\mathcal{X}\times\mathcal{X}\to\mathbb{R}_{\ge0}$ be a local similarity (typically a valid kernel on tokens; e.g., Tanimoto on fingerprints or any PSD similarity defined on monomers).
The Global Alignment (GA) kernel sums contributions over all alignments and can be computed by the DP recursion

\begin{equation}
\label{eq:ga-rec}
\begin{aligned}
M_{0,0}&=1,\qquad M_{i,0}=0,\ M_{0,j}=0,\quad i\ge1,\ j\ge1,\\
M_{i,j}&=\kappa(a_i,b_j)\,\big(M_{i-1,j-1}+M_{i-1,j}+M_{i,j-1}\big),
\end{aligned}
\end{equation}

and set $K_{\mathrm{GA}}(A,B):=M_{n,m}$. Intuitively, each monotone path contributes the product of local similarities along its matched positions, while horizontal/vertical steps allow gaps. Positive definiteness is guaranteed if the transformed local kernel $\kappa/(1+\kappa)$ is positive definite on $\mathcal{X}$ \citep{cuturi2007kernel,cuturi2011fast}. Furthermore the diagonal-dominance concern reported in \citet{cuturi2007kernel} can be mitigated by appropriate scaling of the local kernel (e.g., the temperature $\lambda$) and by avoiding extreme length disparities \citep{cuturi2011fast}.

To incorporate positional information and reduce cost, \cite{cuturi2011fast} introduced \emph{Triangular Global Alignment} (TGA) kernels where  the local similarity is modulated by a Toeplitz position kernel $\omega(i,j)=\psi(|i-j|)$ with compact support
of width
$T \in \mathbb{N}$, i.e.\ $\omega(i,j)=0$ whenever $|i-j|\ge T$
(triangular/band-limited weighting). Here $T$ is a natural number user-chosen bandwidth (maximum
positional lag) controlling how far from the diagonal matches are allowed.
Using such $\omega$ within GA yields a p.d.\ kernel whose computation drops to $O(T\,\min\{n,m\})$.


\paragraph{What we use in this work.}
We employ the standard GA recursion in defined in Equation~\eqref{eq:ga-rec} with a chemically meaningful local kernel at the monomer level (Section~\ref{sec:m-gak}), and use cosine (unit-diagonal) normalization of the resulting Gram matrix for GP inference. 

\[
  \hat{K}(A,B)\;=\;\frac{K_{\mathrm{GA}}(A,B)}{\sqrt{K_{\mathrm{GA}}(A,A)\,K_{\mathrm{GA}}(B,B)}}.
\]


\subsection{Monomer Decoupled Global Alignment Kernel}\label{sec:md-gak}

\paragraph{Representation.}
Each cyclic peptide $M$ is represented as an ordered sequence of monomer SMILES $(s_1,\ldots,s_n)$ extracted from CycPeptMPDB \citep{li2023cycpeptmpdb}. For every monomer we compute
a count-based Morgan fingerprint (radius 3, chirality) using RDKit’s GetMorganFingerprint, represented as a sparse, hashed feature vector. 
This encodes local chemical neighborhoods at the monomer level while preserving sequence order, a crucial  factor for cyclic-peptide permeability \citep{liu2025systematic}.
\paragraph{Local chemical kernel.}
We use the Tanimoto (Jaccard) similarity between monomer fingerprints as the local kernel,
\begin{equation} \label{eq:tanimoto_finger}
\kappa_0\!\big(\phi(s),\phi(t)\big)\;=\;\frac{\langle \phi(s),\phi(t)\rangle}{\|\phi(s)\|_1+\|\phi(t)\|_1-\langle \phi(s),\phi(t)\rangle},
\end{equation}
The Tanimoto kernel on bit/vector fingerprints is positive definite and widely used in cheminformatics \citep{ralaivola2005graph,tripp2023tanimoto}. 
In particular, for nonnegative fingerprint representations the Tanimoto similarity defines a PSD
kernel(\citealp{ralaivola2005graph,tripp2023tanimoto}
for proofs and discussion)
In our default model we set $\kappa=\kappa_0$.

\paragraph{Monomer Decoupled Global Alignment Kernel (MD\text-{}GAK)}
Let $A=(s_1,\ldots,s_n)$ and $B=(t_1,\ldots,t_m)$, two sequences of monomers representing two peptides, respectively. We specialize global alignment to monomer similarities but \emph{decouple} the effect of chemical matches from the effect of gaps. Concretely, let $\kappa(\phi(s_i),\phi(t_j))\in[0,1]$ be the local monomer kernel (Section~\ref{sec:md-gak}), and let $\lambda=1$ be a gap decay. We define the dynamic program
\begin{align}\label{eq:mdgak-dp}
    & M_{0,0}=1,\quad M_{i,0}=M_{0,j}=0,\nonumber &\\
    & M_{i,j}=\kappa\!\big(\phi(s_i),\phi(t_j)\big)\,M_{i-1,j-1}+\lambda\,M_{i-1,j}+\lambda\,M_{i,j-1}, &
\end{align}

and set the kernel value to $K_{\mathrm{MD\text-{}GAK}}(A,B):=M_{n,m}$. Finally, by using cosine normalization we scale $K_{\mathrm{MD\text-{}GAK}}$ in $[0, 1]$.

\[
  \hat{K}_{\mathrm{MD\text-{}GAK}}(A,B)\;=\;\frac{K_{\mathrm{MD\text-{}GAK}}(A,B)}{\sqrt{K_{\mathrm{MD\text-{}GAK}}(A,A)\,K_{\mathrm{MD\text-{}GAK}}(B,B)}}.
\]

\paragraph{Why the decoupling helps.}
In the canonical GA update
\[
M_{i,j}
=
\kappa\big(\phi(s_i),\phi(t_j)\big)\,
\big(M_{i-1,j-1}+M_{i-1,j}+M_{i,j-1}\big),
\]
the local similarity $\kappa(\phi(s_i),\phi(t_j))$ {multiplicatively} controls
all three transitions into $(i,j)$. A poor local match ($\kappa\approx 0$) drives
$M_{i,j}$ close to zero regardless of how good the predecessor states are, so any
alignment path that must pass through $(i,j)$ is heavily downweighted. This can
make the kernel overly sensitive to isolated mismatches and encourage reliance
on a few high-$\kappa$ matches.

In contrast, the decoupled MD-GAK recursion in~\eqref{eq:mdgak-dp} separates the
roles of matches and gaps. If $\kappa\big(\phi(s_i),\phi(t_j)\big)=0$, then
\[
M_{i,j} = M_{i-1,j} + M_{i,j-1},
\]
so the dynamic program can simply bypass the mismatch at $(i,j)$ through gap
steps, with their cost determined solely by $M_{i-1,j}$ and $M_{i,j-1}$.
More generally, $\kappa\big(\phi(s_i),\phi(t_j)\big)$ affects only the diagonal
(match) transition, while insertions and deletions are controlled independently
(by $\lambda=1$ in our setting). This encodes the inductive bias that
chemical similarity belongs to matches, and penalties belong to gaps.

A practical consequence is improved robustness: MD-GAK is less sensitive to single mismatches, and long gap runs do not accumulate large products of local similarities. 
The runtime remains $O(nm)$ per pair. In the next section, we show that the decoupled global alignment still defines a valid positive semidefinite kernel.

\begin{theorem}[Positive semidefiniteness of the Decoupled GA kernel]
\label{prop:dgak-psd}
Let $\mathcal{X}$ be the set of monomers and let $k:\mathcal{X}\times\mathcal{X}\to\mathbb{R}_{\ge 0}$ be a positive semidefinite (PSD) local kernel (e.g., Tanimoto on Morgan fingerprints). Fix $\lambda = 1$. For sequences $A=(s_1,\dots,s_n)$ and $B=(t_1,\dots,t_m)$, define the dynamic program
\[
M_{0,0}=1,\qquad M_{i,0}=M_{0,j}=0,\qquad
M_{i,j}=k(s_i,t_j)\,M_{i-1,j-1}+M_{i-1,j}+M_{i,j-1},
\]
and set $K(A,B):=M_{n,m}$. Then $K$ is a PSD kernel on the space of finite monomer sequences.
\end{theorem}

\noindent The proof is provided in Appendix~\ref{append:proof}.

\paragraph{Position-aware MD-GAK (PMD-GAK).}
To encode positional priors and obtain banded computation, we modulate the local similarity by (i) a soft-match transform and (ii) a compactly supported Toeplitz position kernel (the \emph{triangular} window) as in Triangular GA \cite{cuturi2011fast}. 
We take the Tanimoto kernel
$\kappa_0$ on Morgan fingerprints introduced above as a base monomer kernel, which satisfies
$\kappa_0\in[0,1]$, and define the distance-like score
$\varphi=1-\kappa_0$.
For \(\beta>0\) we use the soft local kernel
\[
  \kappa_\beta\!\big(\phi(s_i),\phi(t_j)\big)
  \;=\;
  \exp\!\Big(-\beta\,\big[1-\kappa_0\!\big(\phi(s_i),\phi(t_j)\big)\big]\Big).
\]

\begin{figure}[ht]
\centering
  \includegraphics[width=\linewidth]{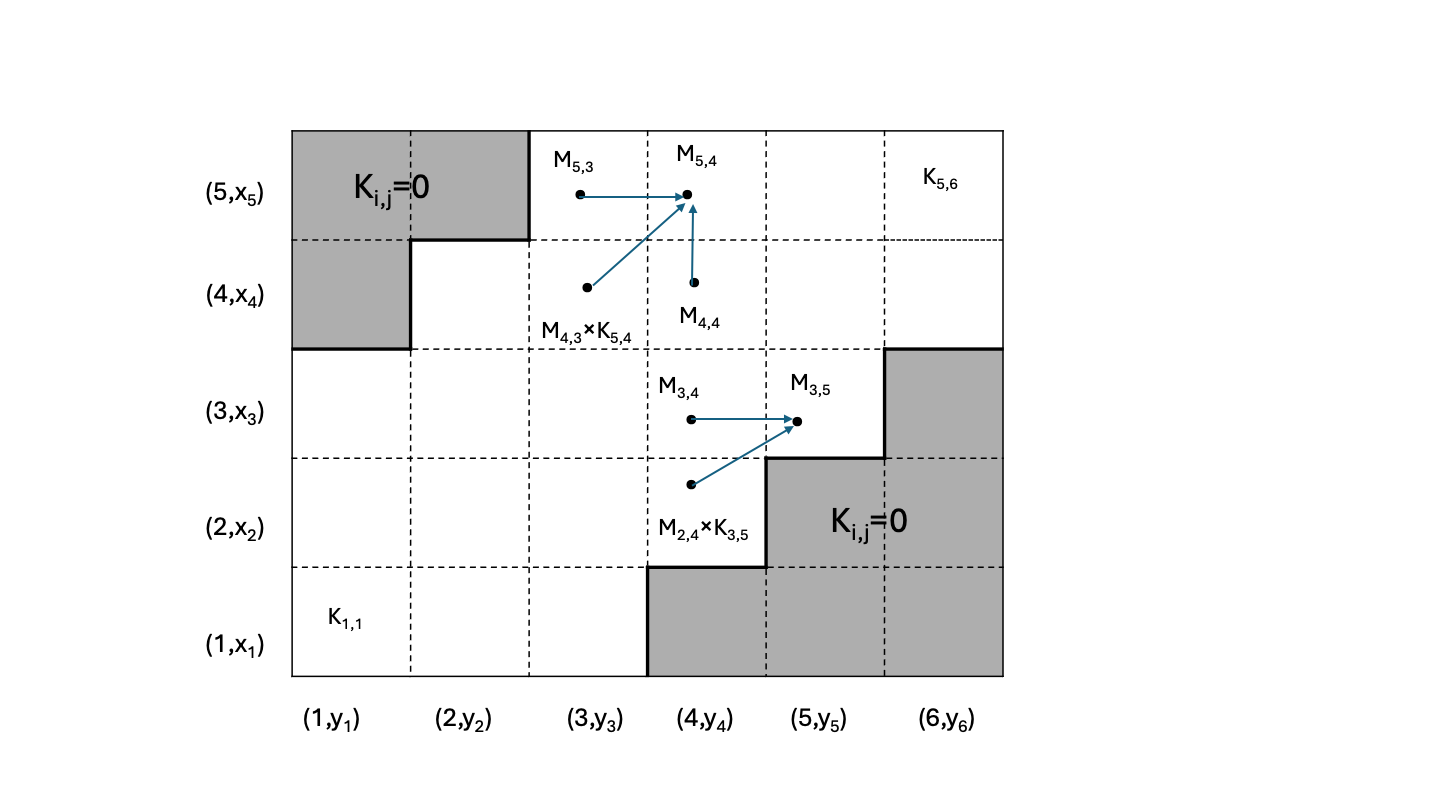}
\caption{Illustration of the PMD-GAK dynamic program with a compactly supported
Toeplitz position kernel $\omega_T(i,j)=\psi(|i-j|)$ of bandwidth $T=3$.
Gray cells indicate entries where $\omega_T(i,j)=0$ (hence
$\kappa_T(i,j)=0$), so $M_{i,j}$ does not need to be updated. For white
cells inside the band, $M_{i,j}$ is computed from its three predecessors
according to
$M_{i,j}=\kappa_T(i,j)M_{i-1,j-1}+M_{i-1,j}+M_{i,j-1}$ (arrows). Because
$\omega_T(i,j)$ depends only on the index difference $|i-j|$ (Toeplitz
structure), the nonzero entries form a diagonal band around the main
diagonal.
}
\label{fig:toeplitz}
\end{figure}

Let the triangular Toeplitz position kernel of bandwidth \(T\in\mathbb{N}\) be
\[
  \omega_T(i,j)\;=\;\max\!\Big\{0,\;1-\frac{|i-j|}{T}\Big\},\qquad
  \omega_T(i,j)=0\ \text{if }|i-j|>T.
\]
Our position-aware local kernel is \(\kappa_T(i,j)=\omega_T(i,j)\,\kappa_\beta(\phi(s_i),\phi(t_j))\), and the DP becomes

\begin{equation}
\label{eq:pmgak}
M_{0,0}=1,\; M_{i,0}=M_{0,j}=0,\qquad
M_{i,j}=\kappa_T(i,j)\,M_{i-1,j-1}+M_{i-1,j}+M_{i,j-1},
\end{equation}

with \(K_{\mathrm{PMD\text-{}GAK}}(A,B)=M_{n,m}\) and optional cosine normalization \({\hat{K}}_{\mathrm{PMD\text-{}GAK}}(A,B)=M_{n,m}\).
Because the triangular window \(\omega_T\) is Toeplitz and compactly supported, the computation is restricted to the diagonal band \(|i-j|\le T\), visiting \(O\!\big(T(n{+}m)\big)\) cells in practice (Fig. \ref{fig:toeplitz}). 
Under the same sufficient conditions used for global-alignment kernels—namely, a suitably transformed positive-definite local kernel and a positive-definite positional kernel—the position-weighted similarity remains positive definite.

\begin{theorem}[Positive semidefiniteness of the Position-aware Decoupled GA kernel]
\label{prop:pmgak-psd}
Let $\mathcal{X}$ be the set of monomers and let $\kappa_0:\mathcal{X}\times\mathcal{X}\to[0,1]$ be a positive semidefinite (PSD) local kernel (e.g., Tanimoto on Morgan fingerprints). Define the distance-like score $\varphi(x,y)=1-\kappa_0(x,y)$ and, for $\beta>0$, the soft local kernel
\[
  \kappa_\beta\!\big(\phi(s_i),\phi(t_j)\big)
  \;=\;
  \exp\!\Big(-\beta\,\big[1-\kappa_0\!\big(\phi(s_i),\phi(t_j)\big)\big]\Big).
\]
Let the triangular Toeplitz position kernel of bandwidth $T\in\mathbb{N}$ be
\[
  \omega_T(i,j)\;=\;\max\!\Big\{0,\;1-\frac{|i-j|}{T}\Big\},\qquad
  \omega_T(i,j)=0\ \text{if }|i-j|>T.
\]
Define the position-aware local kernel
\[
  \kappa_T(i,j)\;=\;\omega_T(i,j)\,\kappa_\beta\big(\phi(s_i),\phi(t_j)\big).
\]
For sequences $A=(s_1,\dots,s_n)$ and $B=(t_1,\dots,t_m)$, define the dynamic program
\[
M_{0,0}=1,\qquad M_{i,0}=M_{0,j}=0,\qquad
M_{i,j}=\kappa_T(i,j)\,M_{i-1,j-1}+M_{i-1,j}+M_{i,j-1},
\]
and set \(K_{\mathrm{PMD\text-{}GAK}}(A,B)=M_{n,m}\). Then $K_T$ is a PSD kernel on the space of finite monomer sequences.
\end{theorem}

\noindent The proof is provided in Appendix~\ref{append:pmgak-proof}.

\paragraph{Relation to graph/OT kernels.}
MD-GAK/PMD-GAK align {sequences} of chemically meaningful monomers, whereas recent OT-based graph kernels used with GPs operate directly on molecular (atom--bond) graphs. In particular, the Wasserstein Weisfeiler--Lehman (WWL) kernel and its Sliced-Wasserstein variant (SWWL) embed node attributes via a (continuous) WL scheme and then compare the resulting {distributions} with (sliced) Wasserstein distances, yielding positive-definite graph kernels well-suited to GP regression on large graphs \citep{togninalli2019wasserstein,perez2024gaussian}. 
For cyclic peptides—where local residue chemistry and backbone ordering both matter—monomer-aware alignment provides a complementary inductive bias: local chemistry shapes match scores through $\kappa_0$, while ordering and allowable warps are handled by the alignment DP (optionally modulated by the positional window $\omega_T$). By contrast, WWL/SWWL aggregate over distributions of WL node embeddings and do not \emph{explicitly} align monomer sequences or enforce a global cyclic order (unless such order is encoded as graph attributes or special edge directions) \citep{togninalli2019wasserstein,perez2024gaussian}.

\subsection{Gaussian processes with molecular fingerprint kernels for peptides}
\label{sec:fp-gp}

In small-molecule chemoinformatics, Gaussian processes (GPs) commonly use the
\emph{Tanimoto} (Jaccard) kernel on binary molecular fingerprints (e.g., ECFP/Morgan), and have been deployed for regression, classification, and Bayesian optimisation \cite{griffiths2023gauche,moss2020gaussian,gosnell2024gaussian,tripp2023tanimoto}.  In protein sequence design, recent BO work has also explored GP surrogates with either string or {fingerprint-style} kernels defined on sequence encodings \cite{benjamins2024bayesian}.
However, to our knowledge, applying a molecular {fingerprint} kernel GP directly to model cyclic-peptide permeability, and comparing it head-to-head against peptide-specific alignment kernels (GAK/MD-GAK/PMD-GAK) within the same evaluation protocol, has not been reported prior to this work.

\paragraph{Fingerprint representation and kernel.}
We used the same similarity directly as described in Equation~\ref{eq:tanimoto_finger} at the whole-peptide level. Tanimoto kernels on bit vectors are positive (semi)definite and standard in GP modelling for molecules \cite{moss2020gaussian,tripp2023tanimoto}. 
This yields our GP with \texttt{TAN\_sim} model.

\paragraph{Convex kernel combination.}
To probe complementarity between  our {decoupled global alignment} and {fingerprint} similarity, we also use a convex combination
\begin{equation}
K_{\mathrm{mix}} \;=\; \alpha\,K_{\mathrm{MD-GAK}} \;+\; (1-\alpha)\,K_{\mathrm{TAN}}, \qquad \alpha\in[0,1],
\end{equation}
with $\alpha$ selected by inner validation. This preserves positive definiteness and lets the GP interpolate between alignment-aware and substructure-aware inductive biases.

Fingerprint-kernel GPs are well established for small molecules and Bayesian Optimization (BO) (e.g., GAUCHE’s Tanimoto kernels \cite{griffiths2023gauche}; FlowMO’s GPU Tanimoto GP \cite{moss2020gaussian}; ordinal-chemistry GP using Tanimoto distances \cite{gosnell2024gaussian}; random-feature approximations for Tanimoto \cite{tripp2023tanimoto}).

\subsection{Dataset and data preparation}
We use the CycPeptMPDB  \citep{li2023cycpeptmpdb}, which compiles permeability measurements for \(\sim\)7{,}334 cyclic peptides (sequence lengths 2–15) drawn from dozens of primary sources and multiple assays, including PAMPA, Caco-2, MDCK and RRCK. Permeability values are reported on a logarithmic scale and clipped to \([-10,-4]\) in CycPeptMPDB; peptides with \(P\geq -6\) are generally regarded as cell-permeable \citep{li2023cycpeptmpdb,liu2025systematic}. We evaluate our models in two complementary settings detailed below.

\paragraph{Setting A: applicability-domain–aware splits}
Following the data-handling principles and splitting strategies highlighted by \citet{geylan2024methodology}, we first extract the SMILES, monomer sequences and PAMPA values for all peptides with available PAMPA in CycPeptMPDB (7{,}298 entries after initial parsing). To mitigate data leakage from near-identical structures, we group duplicates by canonical SMILES (chirality retained) using Morgan fingerprint (radius 3); for each group we average reported PAMPA values to obtain a single label, yielding 7{,}221 unique peptides and 276 unique monomers overall. Averaging duplicates is a common choice~\citet{liu2025systematic}.

\emph{Nested cross-validation.}  5 outer folds (80\% train / 20\% test per outer fold) and, within each outer training split, a 5-fold inner CV for model selection. Using the \(P\geq -6\) threshold, the resulting class counts are 4{,}801 non-permeable and 2{,}420 permeable peptides. In line with \citet[Experiment~7]{geylan2024methodology}, we consider two stratification schemes when constructing folds/splits:  

(i) \emph{Label-stratified} (baseline): stratify by the binary permeability label (\(P\geq -6\) positive) to preserve class balance across folds.  
(ii) \emph{Canonical-group–stratified}: stratify by  canonical groups , ensuring all members of a group reside in the same fold to curb leakage from highly similar peptides.  
For this setting, only PAMPA measurements are used (to reduce inter-assay variability), to avoid source-driven leakage and to define a clear applicability domain. We report ACC, F1, ROC–AUC, and Expected Calibration Error (ECE) with \(M{=}30\) equal-width bins to measure miscalibration  gap between predictive confidence and accuracy \cite{naeini2015obtaining, guo2017calibration, chen2023calibrating, liu2020simple}:
\[
\mathbb{E}\big[\Pr(Y=\hat{Y}\mid \hat{P}=p)-p\big],
\]

\paragraph{Setting B: length-focused PAMPA subset and scaffold splits }
To mirror the benchmark protocol of \citet{liu2025systematic}, we narrow the chemical space to the sequence lengths with sufficient coverage (6, 7 and 10) and exclude non-PAMPA assays. This yields 5{,}758 peptides plus 68 peptides with duplicate PAMPA measurements; these duplicates are retained as independent samples but \emph{always} allocated to the training set during splitting to prevent leakage, giving a working subset of 5{,}826 samples \citep{liu2020simple}. We evaluate two splitting strategies:  
(i) \emph{Random split}: 8:1:1 into train/validation/test (repeated 10 times with different seeds), resulting in 4{,}674/576/576 samples per split.  
(ii) \emph{Scaffold split}: generate Murcko scaffolds with RDKit (ignoring chirality), sort scaffolds by frequency, assign the most frequent to training and the most diverse to test within each sequence-length bucket, then merge to an overall 8:1:1 split (4{,}721/554/551) \citep[Methods]{liu2025systematic}.  
Binary labels use the same threshold \(P\geq -6\) (\(1\) = permeable, \(0\) = non-permeable). As in \citet{liu2025systematic}, this split design probes generalization both under i.i.d. (\emph{random}) and distribution-shifted (\emph{scaffold}) regimes. We report ROC-AUC Score, aligned with the benchmarks.

\subsection{Model implementation}
\paragraph{Baseline models}
We used 2048-bit Morgan fingerprints with radius 3 and total counts for representing peptides smiles to the models which is a robust choice for peptide representations. We trained, GP with GAK \cite{cuturi2011fast}, Random Forest \cite{breiman2001random}, XGBoost \cite{chen2016xgboost} and fine-tuned ChemBerta \cite{chithrananda2020chemberta} as baseline models for the applicability domain aware setting. We tuned models in inner loops validation split and used the best hyper parameters on validation set and reported the model performance on the test results. in setting B, We compared the results on pre-specified benchmarks in \cite{liu2025systematic} including RF \cite{breiman2001random}, SVM \cite{cortes1995support}, AttentiveFP \citep{xiong2019pushing}, DMPNN \citep{yang2019analyzing}, GAT \cite{velivckovic2017graph}, GCN \cite{kipf2016semi}, MPNN \citep{gilmer2017neural}, PAGTN \cite{chen2019path}, RNN \cite{elman1990finding}, LSTM \cite{hochreiter1997long}, GRU \cite{ravanelli2018light}, ChemCeption \cite{goh2017chemception}, ImageMol \cite{zeng2022accurate}, Multi-CycGT \cite{cao2024multi_cycgt} and MUCoCP \cite{yu2024mucocp}.

\section{Results}
\subsection{Setting A: applicability-domain–aware splits}
We evaluate two leakage-controlled protocols: (i) \textbf{label-stratified} splits and (ii) \textbf{canonical-group–stratified} splits that remove near-duplicates. Tables~\ref{tab:label-stratified_alignment} and \ref{tab:canonical-group-stratified_alignment} report results for alignment-aware GPs (GAK/MD-GAK/PMD-GAK) versus strong vector baselines (XGBoost, RF) and a transformer language model (ChemBERTa). Tables~\ref{tab:label-stratified_tanimoto} and \ref{tab:canonical-group-stratified_tanimoto} extend the comparison to {fingerprint} kernels: a GP with Tanimoto similarity ({TAN\_sim}) and its convex combination with DGAK. All metrics are reported as mean~$\pm$~s.e.m. over outer folds; \emph{all values are scaled by $100$} for readability.

\paragraph{Label-stratified split (alignment vs.\ vector baselines).}
Under label stratification (Table~\ref{tab:label-stratified_alignment}), the {GP with MD-GAK} achieves the best threshold metrics---{ACC} (83.0~$\pm$~0.5) and {F1} (73.7~$\pm$~0.9)---while {PMD-GAK} attains the lowest \textbf{Brier} (13.83~$\pm$~0.56) and the best {ECE} among GPs (9.94~$\pm$~1.85), indicating stronger calibration. Tree ensembles display competitive calibration (ECE 8.88/9.07) but \emph{lower discrimination} (ROC–AUC 86.3/85.7) relative to alignment GPs.

\begin{table}[t]
\caption{Label-stratified split (values $\times 100$). Alignment-aware GPs improve threshold metrics and calibration over vector baselines while maintaining strong discrimination.}
\label{tab:label-stratified_alignment}

\centering
\begingroup
\setlength{\tabcolsep}{2.5pt}
\renewcommand{\arraystretch}{1.08}
\begin{tabular}{lccccc}
\hline
Model & ACC & F1 & ROC-AUC & Brier score & ECE \\
\hline
GP with GAK kernel        & 81.1 $\pm$ 1.4 & 69.9 $\pm$ 2.9 & 86.1 $\pm$ 1.3 & 15.93 $\pm$ 0.52 & 15.16 $\pm$ 1.69 \\
\textbf{GP with MD-GAK kernel}     & \textbf{83.0 $\pm$ 0.5} & \textbf{73.7 $\pm$ 0.9} & 87.8 $\pm$ 0.7 & 14.30 $\pm$ 0.46 & 12.31 $\pm$ 1.61 \\
GP with PMD-GAK kernel    & 82.6 $\pm$ 0.7 & 73.2 $\pm$ 0.9 & 87.6 $\pm$ 0.5 & \textbf{13.83 $\pm$ 0.56} & \textbf{9.94 $\pm$ 1.85} \\
XGBoost                   & 78.4 $\pm$ 1.2 & 70.8 $\pm$ 1.1 & 86.3 $\pm$ 0.6 & 14.90 $\pm$ 0.51 & \textbf{8.88 $\pm$ 0.66} \\
RF                        & 78.3 $\pm$ 1.2 & 70.1 $\pm$ 1.2 & 85.7 $\pm$ 0.8 & 15.07 $\pm$ 0.57 & 9.07 $\pm$ 1.60 \\
ChemBERTa                 & 78.6 $\pm$ 0.8 & 69.0 $\pm$ 0.5 & 84.4 $\pm$ 0.9 & 16.03 $\pm$ 0.72 & 11.38 $\pm$ 0.92 \\
\hline
\end{tabular}%

\endgroup

\end{table}

\paragraph{Canonical-group–stratified split (alignment vs.\ vector baselines).}
When stratifying by canonical groups (Table~\ref{tab:canonical-group-stratified_alignment}), the task is harder across the board, consistent with reduced leakage from near-duplicates. {PMD-GAK} attains the best {ACC/F1} (80.3/68.2), the lowest {Brier} (15.42), and the strongest {ROC–AUC} among GPs (84.8), while maintaining competitive ECE. Tree ensembles retain the best ECE (XGBoost 9.42; RF 10.52) but trail the best GP in ROC–AUC.

\begin{table}[t]
\caption{Canonical-group–stratified split (values $\times 100$). PMD-GAK provides the best overall balance of discrimination and calibration under the harder split.}
\label{tab:canonical-group-stratified_alignment}
\centering
\begingroup
\setlength{\tabcolsep}{2.5pt}
\renewcommand{\arraystretch}{1.08}

\begin{tabular}{lccccc}
\hline
Model & ACC & F1 & ROC-AUC & Brier score & ECE \\
\hline
GP with GAK kernel        & 79.0 $\pm$ 1.4 & 65.3 $\pm$ 4.0 & 83.0 $\pm$ 1.1 & 20.53 $\pm$ 0.45 & 22.29 $\pm$ 1.33 \\
GP with MD-GAK kernel     & 80.1 $\pm$ 1.4 & 67.8 $\pm$ 4.4 & 84.7 $\pm$ 1.0 & 15.89 $\pm$ 0.61 & 13.35 $\pm$ 1.52 \\
\textbf{GP with PMD-GAK kernel}    & \textbf{80.3 $\pm$ 1.7} & \textbf{68.2 $\pm$ 3.8} & \textbf{84.8 $\pm$ 1.4} & \textbf{15.42 $\pm$ 0.64} & 11.94 $\pm$ 1.93 \\
XGBoost                   & 76.4 $\pm$ 2.9 & 64.6 $\pm$ 8.5 & 83.9 $\pm$ 1.7 & 15.73 $\pm$ 1.02 & \textbf{9.42 $\pm$ 1.48} \\
RF                        & 75.2 $\pm$ 2.9 & 62.4 $\pm$ 6.4 & 81.8 $\pm$ 2.4 & 16.59 $\pm$ 1.32 & 10.52 $\pm$ 2.82 \\
ChemBERTa                 & 75.3 $\pm$ 1.9 & 62.1 $\pm$ 4.1 & 80.0 $\pm$ 0.9 & 18.43 $\pm$ 0.97 & 13.78 $\pm$ 0.82 \\
\hline
\end{tabular}%

\endgroup

\end{table}

\begin{figure*}[h] 
  \centering
  \setlength{\tabcolsep}{1pt} 
  \begin{subfigure}[t]{0.24\textwidth}
    \includegraphics[width=\linewidth,trim={9.5mm 8.5mm 0 0},clip]{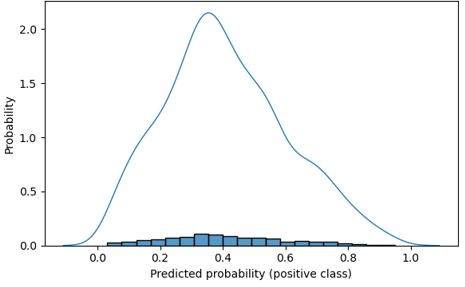}
    \caption{DGAK}
  \end{subfigure}\hfill
  \begin{subfigure}[t]{0.24\textwidth}
    \includegraphics[width=\linewidth,trim={6mm 6.8mm 0 .5mm},clip]{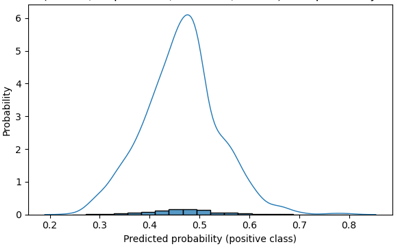}
    \caption{GAK}
  \end{subfigure}\hfill
  \begin{subfigure}[t]{0.24\textwidth}
    \includegraphics[width=\linewidth,trim={9.5mm 6.4mm 0 0},clip]{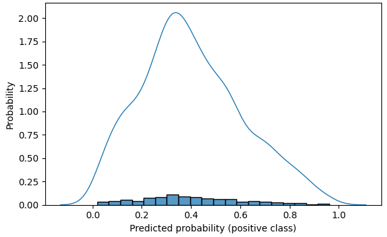}
    \caption{Position DGAK}
  \end{subfigure}\hfill
  \begin{subfigure}[t]{0.24\textwidth}
    \includegraphics[width=\linewidth,trim={9mm 8mm 0 0},clip]{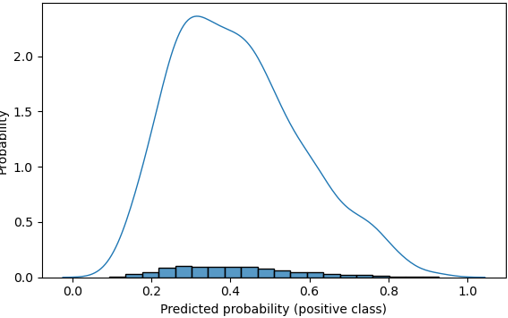}
    \caption{Tanimoto kernel}
  \end{subfigure}\hfill
  \\
  \begin{subfigure}[t]{0.24\textwidth}
    \includegraphics[width=\linewidth,trim={5.3mm 4.5mm 0 .7mm},clip]{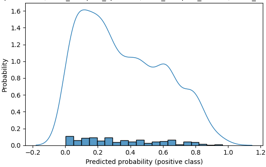}
    \caption{RF}
  \end{subfigure}\hfill
  \begin{subfigure}[t]{0.24\textwidth}
    \includegraphics[width=\linewidth,trim={6.9mm 5.5mm 0 .7mm},clip]{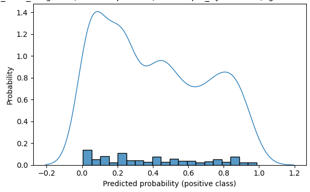}
    \caption{Xgboost}
  \end{subfigure}\hfill
  \begin{subfigure}[t]{0.24\textwidth}
    \includegraphics[width=\linewidth,trim={7.6mm 5.5mm 0 2.5mm},clip]{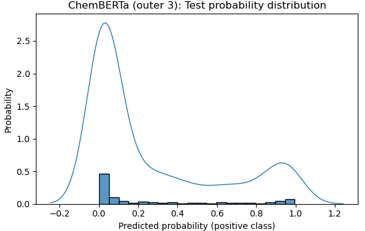}
    \caption{ChemBERTa}
  \end{subfigure}
  \begin{subfigure}[t]{0.24\textwidth}
    \includegraphics[width=\linewidth,trim={16.4mm 12.5mm 0 0},clip]{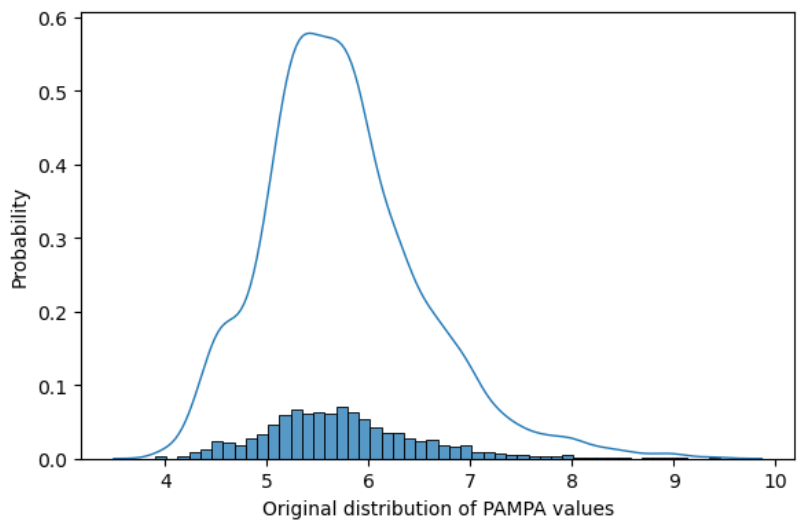}
    \caption{original PAMPA values}
  \end{subfigure}
  \caption{Predicted probability distributions (outer test, canonical-group split). The x-axis shows the predicted probabilities, and the y-axis shows their estimated density of these probabilities on the test set.
  Kernel-based GP models produce score histograms that closely track the empirical distribution of PAMPA values in the dataset, whereas RF, XGBoost, and ChemBERTa yield noticeably different score profiles. This alignment is consistent with their stronger calibration (lower Brier/ECE; Tables \ref{tab:label-stratified_alignment}–\ref{tab:canonical-group-stratified_alignment}) and suggests that the monomer-aware kernels capture permeability-relevant sequence/chemical structure more effectively under canonical splitting.}
  \label{fig:predicted_probs}
\end{figure*}

\paragraph{Score distributions.}
Figure~\ref{fig:predicted_probs} compares model score histograms on the outer test sets under canonical-group stratification. Kernel GPs track the empirical PAMPA distribution more closely (after rescaling), while RF, XGBoost, and ChemBERTa yield skewed or multi-modal profiles. The alignment-aware inductive bias thus provides both improved discrimination and better-behaved probabilistic outputs under the harder split.

\subsection{Setting B: length-focused PAMPA subset (random vs.\ scaffold)}

We further benchmark on a length-focused PAMPA subset containing 6/7/10-mers under two evaluation regimes: a \emph{random} split and a more stringent \emph{scaffold} split. Table~\ref{tab:binary-splits-length} reports ROC–AUC (mean~$\pm$~s.e.m.; values $\times 100$).

The {GP with MD-GAK} attains the strongest discrimination on both splits, reaching \textbf{88.8 $\pm$ 0.2} (random) and \textbf{79.8 $\pm$ 0.0} (scaffold). Compared to the best reported graph models under the same protocols, MD-GAK outperforms \textbf{AttentiveFP} on the random split (86.2 $\pm$ 1.8) and \textbf{MPNN} on the scaffold split (73.4 $\pm$ 8.7). Classical vector baselines (RF/SVM) are substantially weaker on both splits.

\paragraph{Generalization across scaffolds.}
Moving from random to scaffold evaluation reduces performance for all methods, as expected when near-duplicates are removed. The MD-GAK GP exhibits a \emph{9.0}-point drop (88.8 $\rightarrow$ 79.8), which is markedly smaller than the \emph{20.1}-point drop observed for AttentiveFP (86.2 $\rightarrow$ 66.1), suggesting that the alignment-aware inductive bias confers improved robustness to scaffold shifts. MPNN’s scaffold performance is relatively stable but remains below the GP with MD-GAK.

\begin{table}[h]
\caption{Length-focused PAMPA (6/7/10-mers): ROC–AUC under random vs.\ scaffold splits (values $\times 100$).}
\label{tab:binary-splits-length}
\centering
\footnotesize
\setlength{\tabcolsep}{10pt}
\begin{tabular*}{\textwidth}{@{\extracolsep{\fill}}lcc@{}}
\toprule
Method &
\makecell{Random split:\\Classification \textendash{} binary labels} &
\makecell{Scaffold split:\\Classification \textendash{} binary labels} \\
\midrule
\multicolumn{3}{l}{\emph{Alignment-aware GP and classical ML}} \\
\textbf{GP with MD-GAK} & \textbf{$88.8\pm0.2$} & \textbf{$79.8\pm0.0$} \\
RF            & $65.1\pm4.1$          & $55.3\pm3.4$ \\
SVM           & $59.6\pm1.8$          & $53.9\pm0.1$ \\
\addlinespace
\multicolumn{3}{l}{\emph{Graph-based baselines}} \\
\textbf{AttentiveFP (best random)} & \textbf{$86.2\pm1.8$} & \textbf{$66.1\pm5.4$} \\
DMPNN                               & $84.8\pm3.2$          & $71.6\pm7.9$ \\
GAT                                 & $83.9\pm4.4$          & $65.9\pm9.6$ \\
GCN                                 & $77.2\pm5.5$          & $66.9\pm10.4$ \\
\textbf{MPNN (best scaffold)}       & \textbf{$77.2\pm5.5$} & \textbf{$73.4\pm8.7$} \\
PAGTN                               & $78.0\pm6.2$          & $68.2\pm3.2$ \\
\addlinespace
\multicolumn{3}{l}{\emph{String-based baselines}} \\
RNN  & $55.2\pm8.0$ & $52.5\pm3.4$ \\
LSTM & $56.8\pm15.0$ & $53.1\pm10.0$ \\
GRU  & $73.1\pm14.6$ & $61.7\pm13.8$ \\
\addlinespace
\multicolumn{3}{l}{\emph{Image-based baselines}} \\
ChemCeption & $46.6\pm5.1$ & $40.3\pm4.3$ \\
ImageMol    & $80.4\pm1.9$ & $66.1\pm3.7$ \\
\botrule
\end{tabular*}
\end{table}

\subsection{Adding fingerprint kernels: discrimination vs.\ calibration.}
We now introduce peptide-level molecular fingerprints with a {Tanimoto} kernel GP ({TAN\_sim}) and a {convex mixture} with DGAK,(Tables~\ref{tab:label-stratified_tanimoto} and \ref{tab:canonical-group-stratified_tanimoto}). 
Under label stratification, {TAN\_sim} delivers the highest \textbf{ROC–AUC} (\textbf{89.0} $\pm$ 0.8), while MD-GAK remains best on \textbf{ACC/F1} and PMD-GAK remains best calibrated (Brier/ECE). 
Under canonical-group stratification, {TAN\_sim} again yields the top {ROC–AUC} (\textbf{86.7} $\pm$ 0.5). The {convex mixture} achieves the best {ACC/F1} (81.1/69.9), suggesting complementary inductive biases between alignment and substructure similarity. Overall, fingerprint similarity increases \emph{rank discrimination}, whereas alignment-aware kernels improve \emph{probability calibration};

\begin{table}[t]
\caption{Label-stratified split with fingerprint kernels (values $\times 100$). Tanimoto improves rank discrimination; MD/PMD-GAK retain threshold and calibration advantages.}
\label{tab:label-stratified_tanimoto}
\centering
\begingroup
\setlength{\tabcolsep}{2.5pt}
\renewcommand{\arraystretch}{1.08}

\begin{tabular}{lccccc}
\hline
Model & ACC & F1 & ROC-AUC & Brier score & ECE \\
\hline
GP with MD-GAK kernel           & \textbf{83.0 $\pm$ 0.5} & \textbf{73.7 $\pm$ 0.9} & 87.8 $\pm$ 0.7 & 14.30 $\pm$ 0.46 & 12.31 $\pm$ 1.61 \\
GP with PMD-GAK kernel          & 82.6 $\pm$ 0.7 & 73.2 $\pm$ 0.9 & 87.6 $\pm$ 0.5 & \textbf{13.83 $\pm$ 0.56} & \textbf{9.94 $\pm$ 1.85} \\
\texttt{GP with TAN\_sim kernel}   & 82.8 $\pm$ 1.0 & 73.7 $\pm$ 1.4 & \textbf{89.0 $\pm$ 0.8} & 14.40 $\pm$ 0.32 & 14.42 $\pm$ 0.40 \\
GP with convex (DGAK + Tanimoto) & 82.8 $\pm$ 0.7 & 73.6 $\pm$ 1.1 & 88.7 $\pm$ 0.9 & 14.44 $\pm$ 0.33 & 14.28 $\pm$ 0.27 \\
\hline
\end{tabular}%

\endgroup

\end{table}

\begin{table}[t]
\caption{Canonical-group–stratified split with fingerprint kernels (values $\times 100$). Tanimoto again maximizes ROC–AUC; the convex mixture recovers the best ACC/F1 under the harder split.}
\label{tab:canonical-group-stratified_tanimoto}
\centering
\begingroup
\setlength{\tabcolsep}{2.5pt}
\renewcommand{\arraystretch}{1.08}
\begin{tabular}{lccccc}
\hline
Model & ACC & F1 & ROC-AUC & Brier score & ECE \\
\hline
GP with MD-GAK kernel           & 80.1 $\pm$ 1.4 & 67.8 $\pm$ 4.4 & 84.7 $\pm$ 1.0 & 15.89 $\pm$ 0.61 & 13.35 $\pm$ 1.52 \\
GP with PMD-GAK kernel          & 80.3 $\pm$ 1.7 & 68.2 $\pm$ 3.8 & 84.8 $\pm$ 1.4 & \textbf{15.42 $\pm$ 0.64} & \textbf{11.94 $\pm$ 1.93} \\
\texttt{GP with TAN\_sim kernel}   & 80.9 $\pm$ 1.5 & 69.3 $\pm$ 3.1 & \textbf{86.7 $\pm$ 0.5} & 15.57 $\pm$ 0.79 & 13.96 $\pm$ 4.42 \\
GP with convex (DGAK + Tanimoto) & \textbf{81.1 $\pm$ 1.4} & \textbf{69.9 $\pm$ 2.9} & 86.1 $\pm$ 1.3 & 15.93 $\pm$ 0.52 & 15.16 $\pm$ 1.69 \\
\hline
\end{tabular}%

\endgroup

\end{table}

Finally, we assess peptide-level molecular fingerprints on the length-focused subset via a GP with Tanimoto similarity ({TAN\_sim}) and a convex combination with MD-GAK. As shown in Table~\ref{tab:binary-splits_fp}, {TAN\_sim} achieves the highest ROC–AUC on both splits (random: \textbf{0.897 $\pm$ 0.002}; scaffold: \textbf{0.804 $\pm$ 0.000}), narrowly ahead of the MD-GAK GP. The convex mixture (\emph{MD-GAK + Tanimoto}) matches the top ROC–AUC.

\begin{table}[h]
\caption{Length-focused PAMPA (6/7/10-mers): ROC–AUC under random vs.\ scaffold splits using fingerprint kernels (raw values, not scaled).}
\label{tab:binary-splits_fp}
\centering
\footnotesize
\setlength{\tabcolsep}{6pt}
\begin{tabular*}{\textwidth}{@{\extracolsep{\fill}}lcc@{}}
\toprule
Method &
\makecell{Random split:\\Classification \textendash{} binary labels} &
\makecell{Scaffold split:\\Classification \textendash{} binary labels} \\
\midrule
AttentiveFP (best random)                  & $86.2 \pm 1.8$ & $66.1 \pm 5.4$ \\
MPNN (best scaffold)                       & $77.2 \pm 5.5$ & $73.4 \pm 8.7$ \\
{GP with MD-GAK} & {$88.8\pm0.2$} & {$79.8\pm0.0$} \\
\textbf{GP with \texttt{TAN\_sim}} & \textbf{$89.7 \pm 0.2$} & \textbf{$80.4 \pm 0.0$} \\
GP with convex (MD-GAK + Tanimoto)         & $89.7 \pm 0.2$ & $80.4 \pm 0.0$ \\
\botrule
\end{tabular*}
\end{table}

\section{Discussion}

This work aims to reconcile two desiderata for cyclic peptides: a better representation that preserves {ordered monomer chemistry} and a model that remains {data-efficient} with calibrated uncertainties. To this end, we introduced a monomer–aware global alignment family for GPs (MD-GAK and the position-aware PMD-GAK) and evaluated them under applicability-domain–aware protocols that curb leakage from near-duplicates. On CycPeptMPDB, alignment-aware GPs consistently improve discrimination over strong vector and language-model baselines, while PMD-GAK yields the best calibration among GPs (lower Brier/ECE). Under the stricter canonical-group split, the advantage of alignment is amplified (Table~2), and on a length-focused PAMPA subset our GP exceeds, leading graph baselines (Table~5). These findings are robust across splits and metrics, with uncertainty estimates that are well-behaved in the harder setting (Fig.~2).

Although ROC–AUC remains a useful {ranking} metric, permeability datasets are label-imbalanced and downstream decisions hinge on operating at a {single} threshold. In such settings, precision–recall criteria---and in particular the $F_{1}$ score (the harmonic mean of precision and recall)---are more informative than ROC–AUC because ROC can present an overly optimistic picture under class skew \cite{saito2015precision,davis2006relationship}. For this reason, we focus our head-to-head comparison on \textbf{F1} in Tables~\ref{tab:label-stratified_alignment},\ref{tab:canonical-group-stratified_alignment}, under label stratification, {MD-GAK} achieves the best F1 (73.7\,$\pm$\,0.9), and under canonical stratification, {PMD-GAK} leads (68.2\,$\pm$\,3.8), while retaining competitive AUCs {and} improved calibration (lowest Brier/ECE among GPs). This strengthens the practical relevance of alignment kernels for actionable screening, where precision/recall trade-offs matter at deployment time.

\paragraph{Harder tasks magnify the benefit of alignment.}
When we move from label stratification to canonical-group stratification---explicitly removing near-duplicate peptides---performance drops for all methods, but the {relative} improvement of MD-GAK/PMD-GAK over baselines is {stronger}. This is consistent with the hypothesis that alignment captures permeability-relevant {order} and {context} beyond bag-of-substructures; once leakage is reduced, these inductive biases become more valuable (Table \ref{tab:label-stratified_alignment}).

\paragraph{Calibration matters.}
Well-calibrated probabilities enable principled triage in low-data discovery loops. We observe systematically lower Brier/ECE for PMD-GAK relative to other GPs, and tree ensembles remain well calibrated but less discriminative. This mirrors broader evidence that modern ML methods can be poorly calibrated without post-hoc correction \cite{guo2017calibration}. In screening settings with limited assays, these calibrated GP posteriors are advantageous for ranking and for uncertainty-aware decision making. 

\paragraph{Tanimoto at the \emph{monomer} level connects peptides to the small-molecules.}
A key design choice was to score monomer matches with a Tanimoto kernel on Morgan fingerprints inside the alignment DP. This shows that {Tanimoto is effective at the monomer level} for peptides, achieving strong performance. Importantly, this choice bridges peptide modeling with the {mature small-molecule toolkit}: decades of kernels, scalable approximations, and software are immediately compatible. For instance, scalable Tanimoto approximations via random features \cite{tripp2023tanimoto} and comprehensive GP tooling for chemistry (GAUCHE) \cite{griffiths2023gauche} can be dropped in without architectural changes. More broadly, our decoupled setup invites {learned} monomer encoders from chemical language models (e.g., SMILES Transformers, ChemBERTa, MolT5) to provide richer local descriptors within the same GA framework \cite{honda2019smiles,chithrananda2020chemberta,edwards2022translation}. In other words, by proving that "Tanimoto@monomer" is a strong building block, we open a path to upgrade the local kernel $\kappa$ with powerful, pre-trained small-molecule representations as they continue to improve. 

\paragraph{Complementarity of alignment and substructure similarity.}
Across settings, the Tanimoto fingerprint GP (\texttt{TAN\_sim}) consistently maximizes \emph{rank discrimination} (ROC–AUC), whereas alignment variants (MD-/PMD-GAK) dominate {threshold} metrics (ACC/F1) and {calibration} (Brier/ECE). A simple convex kernel (DGAK + Tanimoto) recovers much of both: it matches the strongest AUCs while achieving the best ACC/F1 under the canonical split (Tables~\ref{tab:label-stratified_tanimoto}–\ref{tab:canonical-group-stratified_tanimoto}). These results support a clear division of labor: {local substructures} captured by fingerprints and {ordered monomer context} captured by alignment can encode distinct, complementary signals. In practice, when both high precision/recall and reliable probabilities are required, the convex combination provides a strong, low-complexity default.

\paragraph{Limitations and future work.}
First, we deliberately avoided heavy hyperparameter search; the convex weight $\lambda$ and the positional bandwidth in PMD-GAK were set conservatively. A more systematic selection could further improve threshold metrics per split. Relatedly, moving from single- to multi-task settings (e.g., PAMPA and Caco-2) may benefit from {shared alignment} but {assay-specific} calibration layers.

Second, our local chemistry kernel $\kappa$ used a single family (Tanimoto on Morgan). While effective, bit-vector fingerprints may underrepresent stereochemistry and conformational effects \cite{ucak2023correction,tahil2024stereoisomers}. Future work can (i) upgrade $\kappa$ with richer small-molecule encoders from chemical LLMs (SMILES Transformer, ChemBERTa, MolT5), retaining the same global-alignment scaffold, or (ii) adopt scalable approximations of Tanimoto via random features when data grow \cite{honda2019smiles,chithrananda2020chemberta,edwards2022translation,tripp2023tanimoto}.

Third, scalability: exact GP inference is $\mathcal{O}(N^{3})$ in the number of training peptides. Although our datasets are moderate, larger campaigns will require sparse/inducing-point GPs or Nyström-style approximations \cite{hensman2013gaussian}. These are orthogonal to our kernel design and can be combined with MD-/PMD-GAK and fingerprint kernels without changing the modelling interface.

Fourth, structure granularity. Our approach is \emph{sequence-first}: atom-level arrangements within each monomer and macrocycle are only accessed through the local kernel $\kappa$. Graph GP kernels that capture distributional node/edge structure---for instance, (Wasserstein) Weisfeiler–Lehman variants or topological kernels based on sliced Wasserstein distances---could complement our alignment bias and enable hybrid sequence–graph GPs for macrocycles \cite{togninalli2019wasserstein,perez2024gaussian}. Exploring such hybrids, especially for noncanonical residues and bridged rings, is a promising direction.

Finally, while our results already outperform strong graph baselines on the length-focused subset, more systematic ablations on stereochemistry handling and macrocycle ring closures will clarify where alignment contributes most.

\section{Conclusion}

We introduced monomer-aware global alignment kernels for Gaussian processes (MD-GAK and position-aware PMD-GAK) and showed that they deliver strong discrimination, better calibration, and higher \textbf{F1}—especially under the harder canonical-group split where \textit{alignment matters more}. Complementing alignment with peptide-level molecular fingerprints, a Tanimoto GP (\texttt{TAN\_sim}) maximized AUC, while a simple convex mixture with MD-GAK recovered both top AUC and improved {ACC}/{F1}, confirming the complementarity of local substructures and ordered monomer context. On a length-focused PAMPA subset, our GPs surpassed strong graph baselines and generalized better across scaffolds. Practically, demonstrating that {Tanimoto works at the monomer level} creates a clean interface to the small-molecule ecosystem (classical kernels, scalable random-feature approximations, chemical LLMs) without changing the alignment scaffold, opening an immediate path to richer local chemistry encoders as data scale grows. 

\backmatter

\bmhead{Declarations}
\begin{itemize}
    \item  Funding: This work was supported by a grant from the Swedish Research Council (VR, grant number 2019-00198) as part of the AIR Lund (Artificially Intelligent use of Registers at Lund University) research environment. Additional support was provided by CAISR Health, funded by the Knowledge Foundation (KK-stiftelsen) in Sweden (grant number 20200208 01 H).
    \item Data availability: The data used to train the cell-permeability predictive models for cyclic peptides are publicly available in CycPeptMPDB: \url{http://cycpeptmpdb.com/download/}.
    \item Code availability: The implementation code is available at: \url{https://github.com/ali-amirahmadii/PEPTAK}.
    \item Use of LLMs: LLMs were exclusively used for text cleanup purposes.
    \item Author contributions: AA designed, conceptualized and developed the software and wrote the manuscript. AT designed, conceptualized, and developed the software. All authors were involved in discussions on the project and revised the manuscript. All authors read and approved the final manuscript.
    \item  Conflict of interest: The authors declare no competing interests.
    \item Ethics, consent sections: Not applicable
\end{itemize}

\newpage

\begin{appendices}

\section{Proof}\label{secA1}
\appendix

\subsection{Proof of positive semidefiniteness of the Decoupled GA kernel}\label{append:proof}
Let $s=(s_1,\ldots,s_n)$ and $t=(t_1,\ldots,t_m)$ be two sequences of lengths $n$ and $m$, respectively, where $s_i, t_j \in \mathcal X$. Let $k:\mathcal X\times\mathcal X\to\mathbb{R}$ be a valid kernel that compares elements in the sequences. We define as $M\in\mathbb{R}^{(n+1) \times (m+1)}$ the follow matrix
$$
M(i,j) = k(s_i, t_j)M(i-1,j-1) + M(i-1,j) + M(i, j-1)
$$with initial conditions $M(0,0)=1$, $M(i,0) = 0$, and $M(0, j) = 0$. We want to prove that $M(m,n)$ is valid kernel that compares $s$ and $t$, i.e. $K(s,t) = M(m,n)$.
\\*
\\*
\noindent
A step is one of three moves on the integer lattice:
$R = (1, 0)$, $U =(0, 1)$, and $D=(1,1)$, and a monotone path $\pi$ from $(0,0)$ to $(i,j)$ is a finite sequence of steps that transforms $(0,0)$ to $(i,j)$. For each $\pi$ we collect the set of indices corresponding to diagonal $D$ steps.
$$
D(\pi) = \{\, (i_1, j_1),\ldots,(i_r, j_r), \, i_1<\cdots<i_r, \, j_1 < \cdots <j_r \, \}.
$$
\textbf{Example.} If we consider $\pi$ as follows
$$
\pi:\quad 
(0,0) 
\xrightarrow{\mathrm{R}} (1,0) 
\xrightarrow{\mathrm{D}} (2,1) 
\xrightarrow{\mathrm{U}} (2,2) 
\xrightarrow{\mathrm{D}} (3,3),
$$ $D(\pi) = \{ (2,1),(3,3)\} $.
\\* \\* \noindent We define the weight $w$ of a path 
$$
w(\pi) = \prod_{(i,j) \in D(\pi)} k(s_i,t_j).
$$Note if $\pi$ has no diagonals, the product is empty and equals to 1. Finally we define the set of all the possible monotonic path from $(0,0)$ to $(i,j)$ as $\mathcal P(i,j)$.

\subsection*{Lemma 1}
For each $i,j\ge0$, we want to prove the following
$$
M(i,j) = \sum_{(i,j)\in\mathcal P(i,j)} w(\pi)
$$
\paragraph{Base case.} For $i=j=0$, $\mathcal P(i,j) = \emptyset.$, so $w(\emptyset) = 1$ by definition and follows that $M(0,0) = 1$, which confirms the inital condition on $M(0,0)$.
\paragraph{Indution step.} We assume the equality holds for $i',j'$ such that $i' +j ' < i + j$. Let us know partition $\mathcal P(i,j)$ as follows
\begin{itemize}
    \item Paths whose last step is $R$ (right) are bijection with $\mathcal P(i-1,j)$. It is sufficient to add a $R$ step to $\mathcal P(i-1,j)$ to get $\mathcal P(i,j)$. Note that the bijections holds because $w(\pi)$ depends only on coordinates on $D(\pi)$. The right move $R$ cannot be in $D(\pi)$.
    
    \item Paths whose last step is $U$ (up) are bijection with $\mathcal P(i,j-1)$. 

    \item Paths whose last step is $D$ (diagonal) are 
    $\pi = \pi' \circ D$.
\end{itemize}Putting all together we have
\begin{align}
    &\sum_{\pi \in \mathcal P (i,j)}w(\pi) = \\
    &\sum_{\pi' \in \mathcal P (i-1,j)}w(\pi') + \sum_{\pi' \in \mathcal P (i,j-1)}w(\pi') + w(\{(i,j)\})\sum_{\pi' \in \mathcal P (i-1,j-1)}w(\pi') = \\
    &\sum_{\pi' \in \mathcal P (i-1,j)}w(\pi') + \sum_{\pi' \in \mathcal P (i,j-1)}w(\pi') + k(s_i,t_j)\sum_{\pi' \in \mathcal P (i-1,j-1)}w(\pi').
\end{align}By induction hypothesis since $i-1+j < i+j$, $i+j-1<i+j$, and $i-1+j-1<i+j$, we have that
$$
\sum_{\pi \in \mathcal P (i,j)}w(\pi) = M(i-1,j) + M(i,j-1)+k(s_i,t_j)M(i-1,j-1).
$$

Note that the paths that start with a right or up move lead to prefixes visiting $(i,0)$ or $(0, j)$ with positive index; those prefixes have been given $M(i,0) =0$ and $M(0,j)=0$ The induction sums include these cases but the recurrence's base values ensure they contribute 0, which is consistent with the combinatorial splitting above.

\subsection*{Observation 1}
Since $k$ is valid kernel on sequence elements, it must exist a feature map $\phi$ that maps $\mathcal X$ into a feature vector in $\mathbb{R}^d$, i.e. 
$$
k(s_i, t_j) = \langle\phi(s_i),\phi(t_j)\rangle
$$
Let us take now a diagonal path $\pi \in D(\pi)$, where $\pi = \{ (i_1,j_1),\ldots,(i_r, j_r) \}.$, we define the feature map on $\Phi_{\pi}$ for $s$ and $t$ as
\begin{align}
    \Phi_{\pi}(s) = \phi(s_{i_1}) \otimes \phi(s_{i_2}) \otimes \cdots \otimes \phi(s_{i_r}) \\
    \Phi_{\pi}(t) = \phi(t_{i_1}) \otimes \phi(s_{i_2}) \otimes \cdots \otimes \phi(s_{j_r}).
\end{align}The scalar product between $\Phi_{\pi}(s)$ and $\Phi_{\pi}(t)$ is (by using the properties on Kronecker product $\otimes$)
$$
\langle \Phi_{\pi}(s), \Phi_{\pi}(t) \rangle=\prod_{k=1}^r k(s_{i_k},t_{j_k}) = w(\pi)
$$
\section*{Final proof}
Let us now construct the feature vector $\Psi$ considering all the paths in $\mathcal P(m,n)$ as
$$
\Psi(s) := \bigoplus_{\pi\in\mathcal P(m,n)} \Phi_{\pi}(s).
$$Similarly, we can construct $\Psi(t)$ and their scalar product is 
\begin{align}
    \langle \Psi(s),\Psi(t)\rangle = \langle \bigoplus_{\pi\in\mathcal P(m,n)} \Phi_{\pi}(s),\bigoplus_{\pi\in\mathcal P(m,n)} \Phi_{\pi}(t)\rangle = \sum_{\pi \in \mathcal P(m,n)} \langle\Phi_{\pi}(s),\Phi_{\pi}(t)\rangle = \nonumber \\
    \sum_{\pi \in \mathcal P(m,n)}w(\pi) = M(m,n), \nonumber
\end{align}

which concludes the proof. Since we express $M(m,n)$ as the scalar product of two feature vectors, this is always a valid kernel.

\subsection{Proof of positive semidefiniteness of the Position-aware Decoupled GA kernel}
\label{append:pmgak-proof}

We reuse the notation of Theorem~\ref{prop:dgak-psd}. Recall that $K(A,B)$ in Theorem~\ref{prop:dgak-psd} is PSD for \emph{any} PSD local kernel $k:\mathcal{X}\times\mathcal{X}\to\mathbb{R}_{\ge 0}$. To prove Theorem~\ref{prop:pmgak-psd}, it therefore is enough to show that the position-aware local kernel
\[
  \kappa_T(i,j)\;=\;\omega_T(i,j)\,\kappa_\beta\big(\phi(s_i),\phi(t_j)\big)
\]
is PSD on the extended alphabet of position–monomer pairs. We proceed in three steps.

\subsection*{Step 1: the soft local kernel $\kappa_\beta$ is PSD}

Let $\kappa_0:\mathcal{X}\times\mathcal{X}\to[0,1]$ be a PSD kernel (e.g.\ Tanimoto), and define $\varphi(x,y)=1-\kappa_0(x,y)$. For $\beta>0$ define
\[
  \kappa_\beta(x,y)
  \;=\;
  \exp\big(-\beta\,\varphi(x,y)\big)
  \;=\;
  \exp\big(-\beta\,[1-\kappa_0(x,y)]\big).
\]

Fix a finite set of monomers $\{x_1,\dots,x_N\}\subset\mathcal{X}$ and let
\[
  K_0 \;=\; \big[\kappa_0(x_p,x_q)\big]_{p,q=1}^N.
\]
By assumption $K_0\succeq 0$ and all entries of $K_0$ lie in $[0,1]$. The Gram matrix of $\kappa_\beta$ on this set is
\[
  [K_\beta]_{pq}
  \;=\;
  \kappa_\beta(x_p,x_q)
  \;=\;
  \exp\big(-\beta[1-\kappa_0(x_p,x_q)]\big)
  \;=\;
  e^{-\beta}\,\exp\big(\beta\,\kappa_0(x_p,x_q)\big).
\]

Using the power-series expansion of the exponential,
\[
  \exp(\beta z) = \sum_{m=0}^\infty \frac{\beta^m}{m!}\,z^m,\qquad z\in[0,1],
\]
we obtain the elementwise expansion
\[
  K_\beta
  \;=\;
  e^{-\beta}
  \sum_{m=0}^\infty \frac{\beta^m}{m!}\,K_0^{\ m},
\]
where $K_0^{m}$ denotes the $m$-th Hadamard (entrywise) power: $(K_0^{m})_{pq} = (K_0)_{pq}^m$.

Because $K_0\succeq 0$ and has nonnegative entries, the Schur product theorem implies that each Hadamard power $K_0^{m}$ is PSD. All coefficients $e^{-\beta}\beta^m/m!$ are nonnegative, hence $K_\beta$ is a nonnegative linear combination of PSD matrices and is therefore PSD. As this holds for every finite subset, $\kappa_\beta$ is a positive semidefinite kernel on $\mathcal{X}$.

\subsection*{Step 2: the triangular position kernel $\omega_T$ is PSD}

Fix $T\in\mathbb{N}$. We follow Cuturi~\cite{cuturi2011fast} and use the
triangular Toeplitz position kernel on indices $i,j\in\mathbb{Z}$,
\[
  \omega_T(i,j)
  \;=\;
  \max\Big\{0,\,1-\frac{|i-j|}{T}\Big\},\qquad
  \omega_T(i,j)=0\ \text{if }|i-j|>T.
\]
This is the restriction to $\mathbb{Z}$ of the classical triangular
kernel on $\mathbb{R}$, which is known to be positive definite (see Gneiting~\cite{gneiting2002compactly}). Cuturi~\cite[Sec.~4.3]{cuturi2011fast}
uses the same kernel in the construction of Triangular Global Alignment
(TGA) kernels. It follows that $\omega_T$ is a positive semidefinite
kernel on positions $i,j\in\mathbb{N}$.

\subsection*{Step 3: the position-aware local kernel $\kappa_T$ is PSD}

Consider the product space of positions and monomers
\[
  \widetilde{\mathcal{X}} = \mathbb{Z}\times\mathcal{X},
\]
and define
\[
  k_{\mathrm{loc}}\big((i,s),(j,t)\big)
  := \omega_T(i,j)\,\kappa_\beta(s,t).
\]
Let $\{(i_p,s_p)\}_{p=1}^N \subset \widetilde{\mathcal{X}}$ be arbitrary, and denote
$A_{pq} = \omega_T(i_p,i_q)$ and $B_{pq} = \kappa_\beta(s_p,s_q)$. By Steps~1--2,
$A$ and $B$ are PSD Gram matrices. The Gram matrix of $k_{\mathrm{loc}}$ on these
points is
\[
  G_{pq}
  = k_{\mathrm{loc}}\big((i_p,s_p),(i_q,s_q)\big)
  = A_{pq} B_{pq}
  = (A \circ B)_{pq},
\]
the Hadamard product of $A$ and $B$. By the Schur product theorem, the Hadamard
product of two PSD matrices is PSD, hence $G \succeq 0$. Therefore
$k_{\mathrm{loc}}$ is a PSD kernel on $\widetilde{\mathcal{X}}$.

\subsection*{Step 4: applying the MD-GAK result}

Given a monomer sequence $A=(s_1,\dots,s_n)$, we associate to it the sequence of position–monomer pairs
\[
  \widetilde{A} = \big((1,s_1),\dots,(n,s_n)\big)\in\widetilde{\mathcal{X}}^{\ast},
\]
and similarly for $B=(t_1,\dots,t_m)$.

The dynamic program in~\eqref{eq:pmgak} coincides with the MD-GAK recursion of Theorem~\ref{prop:dgak-psd} applied to the sequences $\widetilde{A},\widetilde{B}$ over the alphabet $\widetilde{\mathcal{X}}$, with local kernel $k_{\text{loc}}$. Since $k_{\text{loc}}$ is PSD, Theorem~\ref{prop:dgak-psd} implies that the resulting sequence kernel
\[
  K_T(A,B)
  \;=\;
  M_{n,m}
\]
is positive semidefinite on the space of finite monomer sequences. This completes the proof of Theorem~\ref{prop:pmgak-psd}.

\newpage

\end{appendices}
\bibliography{sn-bibliography}

@inproceedings{perez2024gaussian,
  title={Gaussian process regression with Sliced Wasserstein Weisfeiler-Lehman graph kernels},
  author={Perez, Rapha{\"e}l Carpintero and Da Veiga, S{\'e}bastien and Garnier, Josselin and Staber, Brian},
  booktitle={International Conference on Artificial Intelligence and Statistics},
  pages={1297--1305},
  year={2024},
  organization={PMLR}
}

@book{williams2006gaussian,
  title={Gaussian processes for machine learning},
  author={Williams, Christopher KI and Rasmussen, Carl Edward},
  volume={2},
  number={3},
  year={2006},
  publisher={MIT press Cambridge, MA},
  address = {USA}
}

@book{scholkopf2002learning,
  title={Learning with kernels: support vector machines, regularization, optimization, and beyond},
  author={Sch{\"o}lkopf, Bernhard and Smola, Alexander J},
  year={2002},
  publisher={MIT press},
  address = {USA}
}

@book{shawe2004kernel,
  title={Kernel methods for pattern analysis},
  author={Shawe-Taylor, John and Cristianini, Nello},
  year={2004},
  publisher={Cambridge university press},
  address   = {Cambridge, UK}
}

@inproceedings{cuturi2011fast,
  title={Fast global alignment kernels},
  author={Cuturi, Marco},
  booktitle={Proceedings of the 28th international conference on machine learning (ICML-11)},
  pages={929--936},
  year={2011}
}

@article{liu2025systematic,
  title={Systematic benchmarking of 13 AI methods for predicting cyclic peptide membrane permeability},
  author={Liu, Wei and Li, Jianguo and Verma, Chandra S and Lee, Hwee Kuan},
  journal={Journal of Cheminformatics},
  volume={17},
  number={1},
  pages={1--12},
  year={2025},
  publisher={Springer}
}

@article{landrum2016rdkit,
  title={Rdkit: open-source cheminformatics http://www. rdkit. org},
  author={Landrum, G},
  journal={Google Scholar There is no corresponding record for this reference},
  volume={3},
  number={8},
  year={2016}
}

@inproceedings{cuturi2007kernel,
  title={A kernel for time series based on global alignments},
  author={Cuturi, Marco and Vert, Jean-Philippe and Birkenes, Oystein and Matsui, Tomoko},
  booktitle={2007 IEEE International Conference on Acoustics, Speech and Signal Processing-ICASSP'07},
  volume={2},
  pages={II--413},
  year={2007},
  organization={IEEE}
}

@article{ralaivola2005graph,
  title={Graph kernels for chemical informatics},
  author={Ralaivola, Liva and Swamidass, Sanjay J and Saigo, Hiroto and Baldi, Pierre},
  journal={Neural networks},
  volume={18},
  number={8},
  pages={1093--1110},
  year={2005},
  publisher={Elsevier}
}

@article{li2023cycpeptmpdb,
  title={CycPeptMPDB: a comprehensive database of membrane permeability of cyclic peptides},
  author={Li, Jianan and Yanagisawa, Keisuke and Sugita, Masatake and Fujie, Takuya and Ohue, Masahito and Akiyama, Yutaka},
  journal={Journal of Chemical Information and Modeling},
  volume={63},
  number={7},
  pages={2240--2250},
  year={2023},
  publisher={ACS Publications}
}

@article{geylan2024methodology,
  title={A methodology to correctly assess the applicability domain of cell membrane permeability predictors for cyclic peptides},
  author={Geylan, G{\"o}k{\c{c}}e and De Maria, Leonardo and Engkvist, Ola and David, Florian and Norinder, Ulf},
  journal={Digital Discovery},
  volume={3},
  number={9},
  pages={1761--1775},
  year={2024},
  publisher={Royal Society of Chemistry}
}

@article{liu2020simple,
  title={Simple and principled uncertainty estimation with deterministic deep learning via distance awareness},
  author={Liu, Jeremiah and Lin, Zi and Padhy, Shreyas and Tran, Dustin and Bedrax Weiss, Tania and Lakshminarayanan, Balaji},
  journal={Advances in neural information processing systems},
  volume={33},
  pages={7498--7512},
  year={2020}
}

@article{chen2023calibrating,
  title={Calibrating transformers via sparse gaussian processes},
  author={Chen, Wenlong and Li, Yingzhen},
  journal={arXiv preprint arXiv:2303.02444},
  year={2023}
}

@inproceedings{guo2017calibration,
  title={On calibration of modern neural networks},
  author={Guo, Chuan and Pleiss, Geoff and Sun, Yu and Weinberger, Kilian Q},
  booktitle={International conference on machine learning},
  pages={1321--1330},
  year={2017},
  organization={PMLR}
}

@inproceedings{naeini2015obtaining,
  title={Obtaining well calibrated probabilities using bayesian binning},
  author={Naeini, Mahdi Pakdaman and Cooper, Gregory and Hauskrecht, Milos},
  booktitle={Proceedings of the AAAI conference on artificial intelligence},
  volume={29},
  number={1},
  year={2015}
}

@incollection{rasmussen2003gaussian,
  title={Gaussian processes in machine learning},
  author={Rasmussen, Carl Edward},
  booktitle={Summer school on machine learning},
  pages={63--71},
  year={2003},
  publisher={Springer},
  address={Berlin, Heidelberg}
}

@article{togninalli2019wasserstein,
  title={Wasserstein weisfeiler-lehman graph kernels},
  author={Togninalli, Matteo and Ghisu, Elisabetta and Llinares-L{\'o}pez, Felipe and Rieck, Bastian and Borgwardt, Karsten},
  journal={Advances in neural information processing systems},
  volume={32},
  year={2019}
}

@article{wang2021effect,
  title={Effect of flexibility, lipophilicity, and the location of polar residues on the passive membrane permeability of a series of cyclic decapeptides},
  author={Wang, Shuzhe and Konig, Gerhard and Roth, Hans-Jorg and Fouch{\'e}, Marianne and Rodde, Stephane and Riniker, Sereina},
  journal={Journal of Medicinal Chemistry},
  volume={64},
  number={17},
  pages={12761--12773},
  year={2021},
  publisher={ACS Publications}
}

@article{hosono2023amide,
  title={Amide-to-ester substitution as a stable alternative to N-methylation for increasing membrane permeability in cyclic peptides},
  author={Hosono, Yuki and Uchida, Satoshi and Shinkai, Moe and Townsend, Chad E and Kelly, Colin N and Naylor, Matthew R and Lee, Hsiau-Wei and Kanamitsu, Kayoko and Ishii, Mayumi and Ueki, Ryosuke and others},
  journal={Nature Communications},
  volume={14},
  number={1},
  pages={1416},
  year={2023},
  publisher={Nature Publishing Group UK London}
}

@article{rogers2010extended,
  title={Extended-connectivity fingerprints},
  author={Rogers, David and Hahn, Mathew},
  journal={Journal of chemical information and modeling},
  volume={50},
  number={5},
  pages={742--754},
  year={2010},
  publisher={ACS Publications}
}

@article{griffiths2023gauche,
  title={GAUCHE: a library for Gaussian processes in chemistry},
  author={Griffiths, Ryan-Rhys and Klarner, Leo and Moss, Henry and Ravuri, Aditya and Truong, Sang and Du, Yuanqi and Stanton, Samuel and Tom, Gary and Rankovic, Bojana and Jamasb, Arian and others},
  journal={Advances in Neural Information Processing Systems},
  volume={36},
  pages={76923--76946},
  year={2023}
}

@article{moss2020gaussian,
  title={Gaussian process molecule property prediction with flowmo},
  author={Moss, Henry B and Griffiths, Ryan-Rhys},
  journal={arXiv preprint arXiv:2010.01118},
  year={2020}
}

@article{gosnell2024gaussian,
  title={A gaussian process model for ordinal data with applications to chemoinformatics},
  author={Gosnell, Arron and Evangelou, Evangelos},
  journal={arXiv preprint arXiv:2405.09989},
  year={2024}
}

@article{tripp2023tanimoto,
  title={Tanimoto random features for scalable molecular machine learning},
  author={Tripp, Austin and Bacallado, Sergio and Singh, Sukriti and Hern{\'a}ndez-Lobato, Jos{\'e} Miguel},
  journal={Advances in Neural Information Processing Systems},
  volume={36},
  pages={33656--33686},
  year={2023}
}

@inproceedings{benjamins2024bayesian,
  title={Bayesian Optimisation for Protein Sequence Design: Gaussian Processes with Zero-Shot Protein Language Model Prior Mean},
  author={Benjamins, Carolin and Surana, Shikha and Bent, Oliver and Lindauer, Marius and Duckworth, Paul},
  booktitle={Machine Learning in Structural Biology Workshop at NeurIPS},
  volume={2024},
  pages={4},
  year={2024}
}

@inproceedings{davis2006relationship,
  title={The relationship between Precision-Recall and ROC curves},
  author={Davis, Jesse and Goadrich, Mark},
  booktitle={Proceedings of the 23rd international conference on Machine learning},
  pages={233--240},
  year={2006}
}

@article{saito2015precision,
  title={The precision-recall plot is more informative than the ROC plot when evaluating binary classifiers on imbalanced datasets},
  author={Saito, Takaya and Rehmsmeier, Marc},
  journal={PloS one},
  volume={10},
  number={3},
  pages={e0118432},
  year={2015},
  publisher={Public Library of Science San Francisco, CA USA}
}

@article{hensman2013gaussian,
  title={Gaussian processes for big data},
  author={Hensman, James and Fusi, Nicolo and Lawrence, Neil D},
  journal={arXiv preprint arXiv:1309.6835},
  year={2013}
}

@article{honda2019smiles,
  title={Smiles transformer: Pre-trained molecular fingerprint for low data drug discovery},
  author={Honda, Shion and Shi, Shoi and Ueda, Hiroki R},
  journal={arXiv preprint arXiv:1911.04738},
  year={2019}
}

@article{chithrananda2020chemberta,
  title={ChemBERTa: large-scale self-supervised pretraining for molecular property prediction},
  author={Chithrananda, Seyone and Grand, Gabriel and Ramsundar, Bharath},
  journal={arXiv preprint arXiv:2010.09885},
  year={2020}
}

@article{edwards2022translation,
  title={Translation between molecules and natural language},
  author={Edwards, Carl and Lai, Tuan and Ros, Kevin and Honke, Garrett and Cho, Kyunghyun and Ji, Heng},
  journal={arXiv preprint arXiv:2204.11817},
  year={2022}
}

@article{ucak2023correction,
  title={Correction: Reconstruction of lossless molecular representations from fingerprints},
  author={Ucak, Umit V and Ashyrmamatov, Islambek and Lee, Juyong},
  journal={Journal of Cheminformatics},
  volume={15},
  number={1},
  pages={68},
  year={2023},
  publisher={Springer}
}

@article{tahil2024stereoisomers,
  title={Stereoisomers are not machine learning’s best friends},
  author={Tah{\i}l, Gökhan and Delorme, Fabien and Le Berre, Daniel and Monflier, {\'E}ric and Sayede, Adlane and Tilloy, S{\'e}bastien},
  journal={Journal of Chemical Information and Modeling},
  volume={64},
  number={14},
  pages={5451--5469},
  year={2024},
  publisher={ACS Publications}
}

@article{senin2008dynamic,
  title={Dynamic time warping algorithm review},
  author={Senin, Pavel},
  journal={Information and Computer Science Department University of Hawaii at Manoa Honolulu, USA},
  volume={855},
  number={1-23},
  pages={40},
  year={2008}
}

@article{meyer2025reverse,
  title={Reverse engineering molecules from fingerprints through deterministic enumeration and generative models},
  author={Meyer, Philippe and Duigou, Thomas and Gricourt, Guillaume and Faulon, Jean-Loup},
  journal={Journal of Cheminformatics},
  volume={17},
  number={1},
  pages={157},
  year={2025},
  publisher={Springer}
}

@inproceedings{gilmer2017neural,
  title={Neural message passing for quantum chemistry},
  author={Gilmer, Justin and Schoenholz, Samuel S and Riley, Patrick F and Vinyals, Oriol and Dahl, George E},
  booktitle={International conference on machine learning},
  pages={1263--1272},
  year={2017},
  organization={Pmlr}
}

@article{yang2019analyzing,
  title={Analyzing learned molecular representations for property prediction},
  author={Yang, Kevin and Swanson, Kyle and Jin, Wengong and Coley, Connor and Eiden, Philipp and Gao, Hua and Guzman-Perez, Angel and Hopper, Timothy and Kelley, Brian and Mathea, Miriam and others},
  journal={Journal of chemical information and modeling},
  volume={59},
  number={8},
  pages={3370--3388},
  year={2019},
  publisher={ACS Publications}
}

@article{xiong2019pushing,
  title={Pushing the boundaries of molecular representation for drug discovery with the graph attention mechanism},
  author={Xiong, Zhaoping and Wang, Dingyan and Liu, Xiaohong and Zhong, Feisheng and Wan, Xiaozhe and Li, Xutong and Li, Zhaojun and Luo, Xiaomin and Chen, Kaixian and Jiang, Hualiang and others},
  journal={Journal of medicinal chemistry},
  volume={63},
  number={16},
  pages={8749--8760},
  year={2019},
  publisher={ACS Publications}
}

@article{gneiting2002compactly,
  title={Compactly supported correlation functions},
  author={Gneiting, Tilmann},
  journal={Journal of Multivariate Analysis},
  volume={83},
  number={2},
  pages={493--508},
  year={2002},
  publisher={Elsevier}
}

@article{breiman2001random,
  title={Random forests},
  author={Breiman, Leo},
  journal={Machine learning},
  volume={45},
  number={1},
  pages={5--32},
  year={2001},
  publisher={Springer}
}

@inproceedings{chen2016xgboost,
  title={Xgboost: A scalable tree boosting system},
  author={Chen, Tianqi and Guestrin, Carlos},
  booktitle={Proceedings of the 22nd acm sigkdd international conference on knowledge discovery and data mining},
  pages={785--794},
  year={2016}
}

@article{cortes1995support,
  title={Support-vector networks},
  author={Cortes, Corinna and Vapnik, Vladimir},
  journal={Machine learning},
  volume={20},
  number={3},
  pages={273--297},
  year={1995},
  publisher={Springer}
}

@article{velivckovic2017graph,
  title={Graph attention networks},
  author={Veli{\v{c}}kovi{\'c}, Petar and Cucurull, Guillem and Casanova, Arantxa and Romero, Adriana and Lio, Pietro and Bengio, Yoshua},
  journal={arXiv preprint arXiv:1710.10903},
  year={2017}
}

@article{kipf2016semi,
  title={Semi-supervised classification with graph convolutional networks},
  author={Kipf, TN},
  journal={arXiv preprint arXiv:1609.02907},
  year={2016}
}

@article{chen2019path,
  title={Path-augmented graph transformer network},
  author={Chen, Benson and Barzilay, Regina and Jaakkola, Tommi},
  journal={arXiv preprint arXiv:1905.12712},
  year={2019}
}

@article{elman1990finding,
  title={Finding structure in time},
  author={Elman, Jeffrey L},
  journal={Cognitive science},
  volume={14},
  number={2},
  pages={179--211},
  year={1990},
  publisher={Wiley Online Library}
}

@article{hochreiter1997long,
  title={Long short-term memory},
  author={Hochreiter, Sepp and Schmidhuber, J{\"u}rgen},
  journal={Neural computation},
  volume={9},
  number={8},
  pages={1735--1780},
  year={1997},
  publisher={MIT press}
}

@article{ravanelli2018light,
  title={Light gated recurrent units for speech recognition},
  author={Ravanelli, Mirco and Brakel, Philemon and Omologo, Maurizio and Bengio, Yoshua},
  journal={IEEE Transactions on Emerging Topics in Computational Intelligence},
  volume={2},
  number={2},
  pages={92--102},
  year={2018},
  publisher={IEEE}
}

@article{goh2017chemception,
  title={Chemception: a deep neural network with minimal chemistry knowledge matches the performance of expert-developed QSAR/QSPR models},
  author={Goh, Garrett B and Siegel, Charles and Vishnu, Abhinav and Hodas, Nathan O and Baker, Nathan},
  journal={arXiv preprint arXiv:1706.06689},
  year={2017}
}

@article{zeng2022accurate,
  title={Accurate prediction of molecular properties and drug targets using a self-supervised image representation learning framework},
  author={Zeng, Xiangxiang and Xiang, Hongxin and Yu, Linhui and Wang, Jianmin and Li, Kenli and Nussinov, Ruth and Cheng, Feixiong},
  journal={Nature Machine Intelligence},
  volume={4},
  number={11},
  pages={1004--1016},
  year={2022},
  publisher={Nature Publishing Group UK London}
}

@article{yu2024mucocp,
  title={MuCoCP: a priori chemical knowledge-based multimodal contrastive learning pre-trained neural network for the prediction of cyclic peptide membrane penetration ability},
  author={Yu, Yunxiang and Gu, Mengyun and Guo, Hai and Deng, Yabo and Chen, Danna and Wang, Jianwei and Wang, Caixia and Liu, Xia and Yan, Wenjin and Huang, Jinqi},
  journal={Bioinformatics},
  volume={40},
  number={8},
  pages={btae473},
  year={2024},
  publisher={Oxford University Press}
}

@article{cao2024multi_cycgt,
  title={Multi\_CycGT: a deep learning-based multimodal model for predicting the membrane permeability of cyclic peptides},
  author={Cao, Lujing and Xu, Zhenyu and Shang, Tianfeng and Zhang, Chengyun and Wu, Xinyi and Wu, Yejian and Zhai, Silong and Zhan, Zhajun and Duan, Hongliang},
  journal={Journal of medicinal chemistry},
  volume={67},
  number={3},
  pages={1888--1899},
  year={2024},
  publisher={ACS Publications}
}


\end{document}